\documentclass[10pt,twocolumn,letterpaper]{article}

\usepackage{cvpr}              % To produce the CAMERA-READY version
\usepackage[dvipsnames]{xcolor}

\usepackage[accsupp]{axessibility} % Improves PDF readability for those with visual impairments.

\definecolor{cvprblue}{rgb}{0.21,0.49,0.74}
\usepackage[pagebackref,breaklinks,colorlinks,citecolor=cvprblue]{hyperref}
\usepackage{parskip}
\usepackage{multirow} % For \multirow command
\usepackage{makecell} % For \makecell command
\usepackage{booktabs} % For \toprule, \midrule, \bottomrule commands
\usepackage{blindtext}% for dummy text only
\usepackage{graphicx}
\usepackage{caption}
\usepackage{siunitx}
\usepackage{verbatim}

\def\paperID{10338} % *** Enter the Paper ID here
\def\confName{CVPR}
\def\confYear{2024}

\title{Segment Every Out-of-Distribution Object}
\author{Wenjie Zhao$^{1}$ \hspace{1cm} Jia Li$^{1}$ \hspace{1cm} Xin Dong$^{2}$ \hspace{1cm} Yu Xiang$^{1}$ \hspace{1cm} Yunhui Guo$^{1}$\\
$^{1}$University of Texas at Dallas \hspace{1cm} $^{2}$ Harvard University\\
$^{1}${\tt\small \{wxz220013, jxl220096, yu.xiang, yunhui.guo\}@utdallas.edu}, $^{2}${\tt\small xindong@g.harvard.edu}}

\begin{document}

%The top row displays real-world images, highlighting anomalous objects with bounding boxes based on training categories. Subsequent rows exhibit the anomaly score maps generated by various methods, including PEBAL\cite{PEBAL}, RPL\cite{RPL}, and our approach, S2M.

\twocolumn[{%
\renewcommand \twocolumn[1][]{#1}%
\maketitle

\begin{center}
    \centering
    \includegraphics[width=\textwidth]{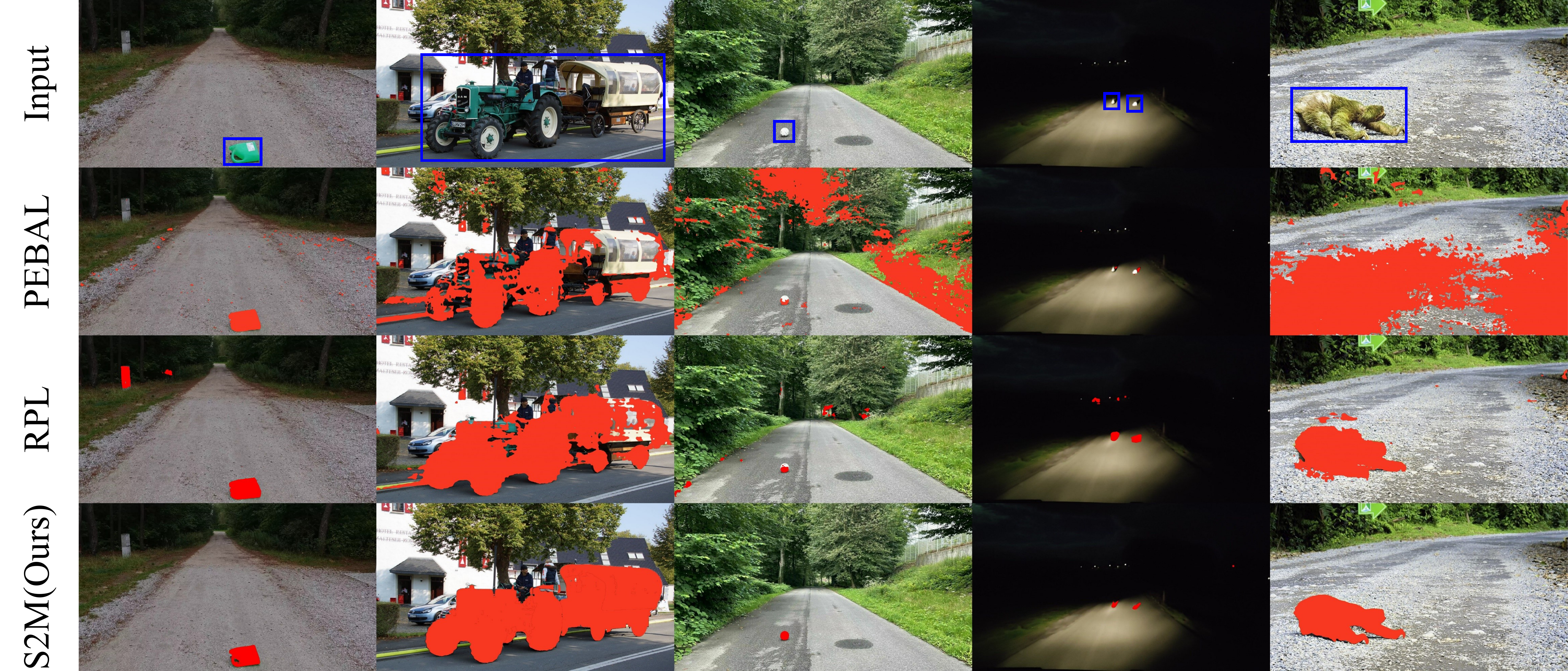}
    %\captionsetup{skip=2pt}  % Adjust the value here
    \captionof{figure}{\textbf{Compared with the state-of-the-art Out-of-Distribution (OoD) detection methods in semantic segmentation, our method excels in producing high-quality masks for OoD objects.} The top row displays several real-world images, highlighting anomalous objects with blue bounding boxes. Subsequent rows present masks generated by different methods for the OoD objects, including PEBAL \cite{PEBAL}, RPL \cite{RPL} and our method S2M. For PEBAL and RPL, the masks are derived from anomaly scores using the optimal threshold specific to each dataset. Unlike other methods that frequently generate noise outside of OoD objects and exhibit fragmented masks, S2M delivers precise masks for the OoD object.}
    \vspace{8pt}
    \label{fig:MainCompare}
\end{center}%
}]

\begin{abstract}
\vspace{-5pt}

Semantic segmentation models, while effective for in-distribution categories, face challenges in real-world deployment due to encountering out-of-distribution (OoD) objects. Detecting these OoD objects is crucial for safety-critical applications. Existing methods rely on anomaly scores, but choosing a suitable threshold for generating masks presents difficulties and can lead to fragmentation and inaccuracy. This paper introduces a method to convert anomaly \underline{S}core \underline{T}o segmentation \underline{M}ask, called S2M, a simple and effective framework for OoD detection in semantic segmentation. Unlike assigning anomaly scores to pixels, S2M directly segments the entire OoD object. By transforming anomaly scores into prompts for a promptable segmentation model, S2M eliminates the need for threshold selection. Extensive experiments demonstrate that S2M outperforms the state-of-the-art by approximately 20\% in IoU and 40\% in mean F1 score, on average, across various benchmarks including Fishyscapes, Segment-Me-If-You-Can, and RoadAnomaly datasets. Code is available at \href{https://github.com/WenjieZhao1/S2M}{https://github.com/WenjieZhao1/S2M}.

\end{abstract}  
\vspace{-5pt}
\section{Introduction}
\label{sec:intro}

Semantic segmentation, a critical task in computer vision, bears substantial significance across various applications such as autonomous driving and aerial imagery analysis \cite{segementImportant,ADriving_Survey}. While current semantic segmentation models demonstrate impressive performance, their practical deployment remains challenging. A significant obstacle lies in their limited ability to detect out-of-distribution (OoD) objects. Specifically, these models often assign pixels within the OoD object to one of the categories used in training, leading to inaccurate segmentation masks \cite{RPL,PEBAL,SAM}. Addressing the OoD detection problem in semantic segmentation is important, given that inaccuracies in OoD objects masks can lead to erroneous conclusions, ultimately posing safety concerns in applications such as autonomous driving  \cite{ADriving_Survey,ADriving_filos20a,ADriving_Nitsch2021,ADriving_Survey2,Fishyscapes}.

Existing OoD detection methods in semantic segmentation address this issue by assigning an anomaly score for each pixel \cite{Npad,RPL,PEBAL,ADpixel1,ADpixel2}. 
%They retrain segmentation model to boosts the uncertainty of OoD pixels \cite{RPL}. 
The pixels with high anomaly scores will be considered as part of OoD objects. The anomaly scores are typically derived by the probabilistic predictions of each pixel made by the segmentation model \cite{Dan_Baseline,standardizedMaxLogits,BayesianSegmentation}. For example, PEBAL \cite{PEBAL} proposes a pixel-wise energy-based segmentation method to compute the anomaly score. RPL \cite{RPL} introduces a residual pattern learning module and computes the anomaly score by an energy-based method \cite{PEBAL}, effectively enhancing the model's sensitivity to OoD pixels without compromising the segmentation performance on in-distribution (ID) data. 

\begin{figure}[t]
    \centering
    \includegraphics[width=1.0\linewidth]{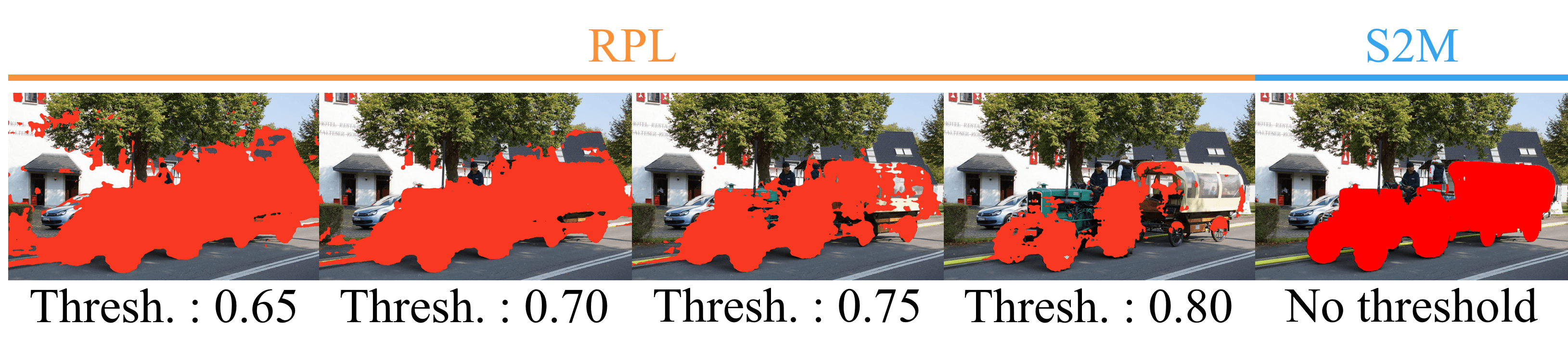}
    \vspace{-22pt}
    \caption{Existing anomaly score-based OoD detection method such as RPL \cite{RPL} is sensitive to the thresholds while \textbf{ S2M eliminates the need for threshold selection}, which is more practical. Besides, S2M also gives a more precise mask. }
    \label{threshold}
\vspace{-5pt}
\end{figure}

While the anomaly score-based OoD detection methods can accurately identify anomalous pixels, they lack an effective way for \emph{segmenting} the entire OoD object. In particular, to derive a mask for OoD objects, it is crucial to use a carefully chosen threshold to distinguish between anomalous and normal pixels~\cite{threshold_Vasiliuk, threshold_cui,threshold_gao}. However, determining the optimal threshold in practical applications can be a difficult task. In Fig. \ref{threshold}, it is evident that the optimal threshold range for existing OoD detection methods is quite narrow; even a slight deviation, either too high or too low, can result in inaccurate segmentation. Choosing the optimal threshold often requires a dedicated validation dataset for fine-tuning the threshold. In practice, such validation datasets are often not available. Even with the optimal threshold, the generated masks from anomaly score-based OoD detection methods may still require refinement, as the anomaly scores for certain pixels might be inaccurate, leading to fragmented or discontinuous masks. Fig. \ref{fig:MainCompare} illustrates various examples of masks generated by PEBAL~\cite{PEBAL} and RPL~\cite{RPL}, two state-of-the-art anomaly score-based OoD detection methods. The fragmented masks can hardly be useful as the model users cannot accurately locate the OoD objects. 

In this paper, we present a method, named S2M, which converts anomaly \underline{S}core \underline{T}o segmentation \underline{M}ask, serving as a simple and general out-of-distribution (OoD) detection framework for semantic segmentation. Specifically, rather than deriving masks for OoD objects through thresholding anomaly scores, S2M aims to \emph{segment} the entire OoD object. Thus S2M effectively mitigates the fragmentation issue associated with employing a threshold. In more details, given any anomaly scores generated by existing OoD detection methods, S2M generates box prompts from them using a prompt generator. The generated box prompt will approximately locate the OoD objects. Then, the box prompts are used as input for a promptable segmentation model to generate the masks for the OoD objects. Fig. \ref{fig:MainCompare} shows several examples of masks generated by S2M for the OoD objects. Different from using a threshold to generate the masks, the masks generated by S2M accurately segment the entire OoD objects while avoiding the inclusion of normal pixels in the mask. Compared with existing OoD detection methods, S2M has several advantages: \textbf{1) Simple:} S2M is a simple pipeline that is easy to train and deploy, requiring no hyperparameter tuning. \textbf{2) General:} S2M can be integrated with any anomaly score-based OoD detection methods to generate high-quality OoD masks. \textbf{3) Effective:} S2M can accurately segment the OoD objects without creating fragmented and inaccurate masks.

%without reducing the segmentation performance of the segmentation model.\wj{NEED REVIEW}

The contributions of our paper are as follows:
\begin{itemize}
    \item We propose S2M, a simple and general pipeline to generate the precise mask for OoD objects.
    \item We eliminate the need to manually choose an optimal threshold for generating segmentation masks, a step that frequently adds complexity to deployment. Additionally, our method is general and independent of particular anomaly scores, prompt generators, or promptable segmentation models.
    
    \item We extensively evaluate S2M on commonly used OoD segmentation benchmarks including Fishyscapes, Segment-Me-If-You-Can, and RoadAnomaly datasets, the results show that S2M improves the state-of-the-art by approximately 20\% in IoU and 40\% in mean F1 score, on average, across all benchmarks.
     
\end{itemize}

\section{Related work}
\label{sec:related}
\textbf{Semantic Segmentation}, a crucial task in computer vision, has seen significant advancements in recent years \cite{SS_Unet,SS_MaskRcnn,SS_RL, SS_transformer,SAM}. Traditionally, this field has been dominated by pixel-wise classification methods, notably influenced by the advent of Fully Convolutional Networks (FCN) \cite{FCN}. These method excel in generating detailed segmentation by preserving high-level image representations and integrating multi-scale contextual information, as evidenced in various architectures and approaches. The DeepLab series, with its use of dilated convolution to enhance the receptive field, represents a notable development in this area. DeepLabv3+ \cite{deeplabv3+} incorporates atrous convolutions to enhance feature extraction, along with a decoder module that refines segmentation results, particularly along object boundaries.

 Recent trends in semantic segmentation have shifted towards transformer-based architectures and attention mechanisms, which offer improved handling of contextual relationships within images \cite{SS_transformer,SS_rethinkingTransformer,SS_Segmenter}. In contrast to these typical segmentation approaches, the prompt-based segmentation presents a unique paradigm. Its primary strength lies in its robust segmentation capabilities, enabling the efficient delineation of objects within an image. For example, CLIPSeg \cite{promptseg} introduced a system capable of generating image segmentation at test time based on any given prompt, whether text or image, thereby enabling a unified model for three distinct segmentation tasks. Segment Anything Model (SAM) \cite{SAM} is known for its robust capability to generate precise masks for every object, demonstrating exceptional performance in object segmentation tasks. 
 %However, unlike conventional models, SAM lack explicit classification abilities. This characteristic, rather than being a limitation, is particularly advantageous in the context of OoD detection. 
 %Therefore, SAM's potent segmentation power can be fully leveraged in OOD detection tasks, capitalizing on its ability to effectively segment without the need for classification.

\textbf{Pixel-Wise OoD detection} has mainly based on the output of the semantic segmentation models. Initially, many OoD detection methods employed mathematical approaches, focusing on the analysis of confidence distributions from segmentation models to identify anomalous pixels \cite{Mahalanobis,generalizedOODsurvey,liang2020enhancing,liu2020energy}. %The Mahalanobis distance is used to detect OoD samples by calculating the distance of a test sample's features from the mean of the training data, considering the covariance of the training data, and flagging samples that exceed a certain threshold as OOD \cite{Mahalanobis}. 
For example, energy-based method \cite{liu2020energy} applies an energy function in place of the softmax function on any model to get pixel-wise anomaly scores without altering the model architecture, and also introduces an energy-based regularization term for targeted fine-tuning of the model. Synboost \cite{Synboost} calculates anomaly scores by ensembling uncertainty maps using learned weights, effectively enhancing anomaly detection while addressing the overconfidence issue. DenseHybrid \cite{Densehybrid} obtains the anomaly score by using a hybrid anomaly detection approach.

In the field of OoD detection in semantic segmentation, a notable shift is occurring where models are being retrained to more effectively heighten their sensitivity in identifying anomalous objects \cite{chen2023lara,dionelis2021few}. Training data are sourced from the Outlier Exposure (OE) process, which involves incorporating certain OoD objects into in-distribution images \cite{OE}, enhancing the model's robustness against anomalies. Despite their roots in semantic segmentation, these existing OoD detection models diverge from traditional segmentation approaches in their output format. Instead of generating masks based on confidence scores, as is common in segmentation, these OoD detection methods produce an anomaly score map as their output. However, using anomaly score maps has practical limitations, as they are less accurate than masks in localizing OoD objects. 

%Recent studies using scoring functions to reduce background noise \cite{RbA} aim for better accuracy, but such probabilistic approaches can be overly complex. An end-to-end pipeline could be a more direct, efficient alternative, suggesting room for improvement in OoD detection methods.

\section{Method}
\begin{figure}[t]
    \centering
    \includegraphics[width=1.0\linewidth]{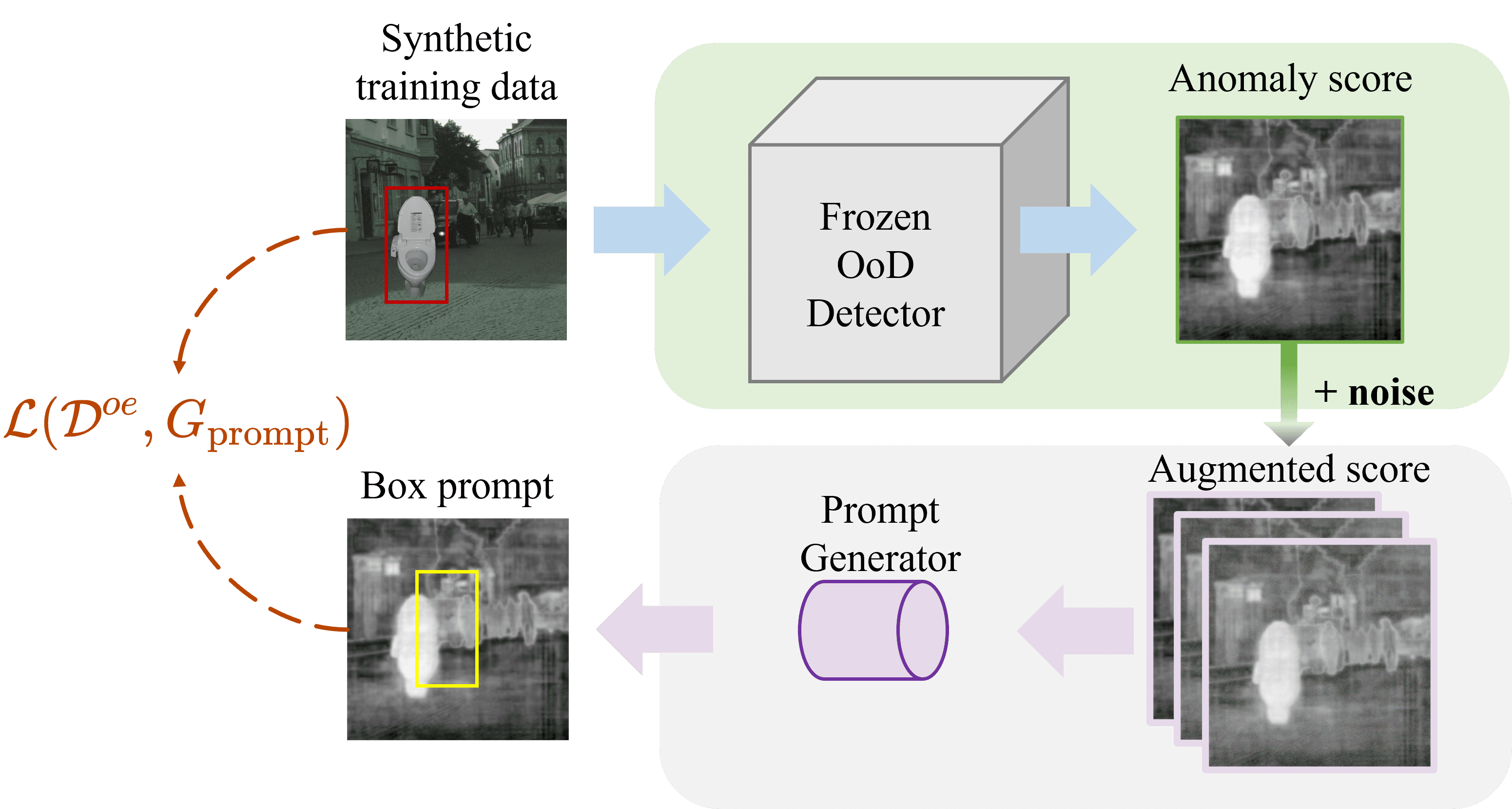}
    \caption{Overview of the training pipeline. We frozen the OoD Detector and only train prompt generator.}
    \label{fig:train}
\vspace{-5pt}
\end{figure}

We tackle the OoD object detection in semantic segmentation with a simple and effective pipeline. Our approach addresses the inherent limitations of existing anomaly score-based OoD detection methods, which mainly provide pixel-wise anomaly scores. While these anomaly scores can indicate whether a given pixel can possibly belong to an OoD object, accurately obtaining the segmentation mask for the entire OoD object is difficult. In contrast, our proposed S2M leverages anomaly score maps to create box prompts that signal the presence of OoD objects. The box prompts serve as input for a promptable segmentation framework, which processes both these prompts and the original image to generate masks for OoD objects. The training pipeline of the framework is depicted in Fig. \ref{fig:train}.

%\textbf{First}, we propose a prompt generator which takes pixel-wise anomalies score map as input and generates box prompts for localizing the OoD object; \textbf{Second}, a synthetic training data is generated for training the prompt generator; \textbf{Finally}, to further enhance the method's performance and reliability, we propose to augment the anomaly scores with additional noises. In the following sections, we will introduce the details of our method.

\subsection{Image to Anomaly Score}

Consider a segmentation network denoted as $f(x;\theta)$ and an input image $x \in \mathbb{R}^{H\times W\times3}$, where $W$ represents the image width and $H$ denotes the image height. The logit for a pixel $i$ produced by the segmentation network can be expressed as $L_i(x;\theta) = (f^1_i(x;\theta), f^2_i(x;\theta), ..., f^C_i(x;\theta) )$, where $C$ represents the total number of classes. $L_i(x;\theta)$ can be normalized using a softmax function as $P_i(x;\theta)$. An anomaly score $S_i(x; \theta)$ can be computed based on the model output. 

One possible anomaly score, computed based on Shannon entropy \cite{Synboost,Gal2016Uncertainty}, is defined as:
\begin{align}
    H_i(x, \theta) = -\sum_{c \in C} P^c_i(x;\theta) \log_{2}P^c_i(x;\theta)
\end{align}
A higher entropy implies that the segmentation model exhibits uncertainty regarding the prediction at pixel $i$, suggesting that pixel $i$ is more likely to belong to an OoD object. Recently, PEBAL \cite{PEBAL} proposed to compute the anomaly score with an energy-based approach,
\begin{equation}
\begin{aligned}
E_i(x;f) &= -T \cdot \sum_{c\in C} e^{f_i^c(x)/T}
\end{aligned}
\end{equation}
Where $T$ is the temperature parameter. As previously discussed, although the anomaly score can identify whether an individual pixel belongs to an OoD object, obtaining the mask for the entire OoD object is challenging.

\subsection{Anomaly Score to Box Prompt}
While directly generating masks from anomaly scores is challenging, the scores themselves still provide a strong indication of the location of the OoD objects. For instance, a region characterized by a majority of pixels with high anomaly scores is likely to contain an OoD object. On the other hand, the emergence of large-scale models in semantic segmentation, equipped to generate accurate segmentation masks with given prompts, signifies a significant recent advancement. For example, the Segment Anything Model (SAM) \cite{SAM}, which can process various prompts to generate segmentation masks. This motivates us to convert anomaly scores into prompts, enabling a promptable segmentation model to generate the masks for the OoD objects. 

\begin{figure}[t]
    \centering
    \includegraphics[width=0.85\linewidth]{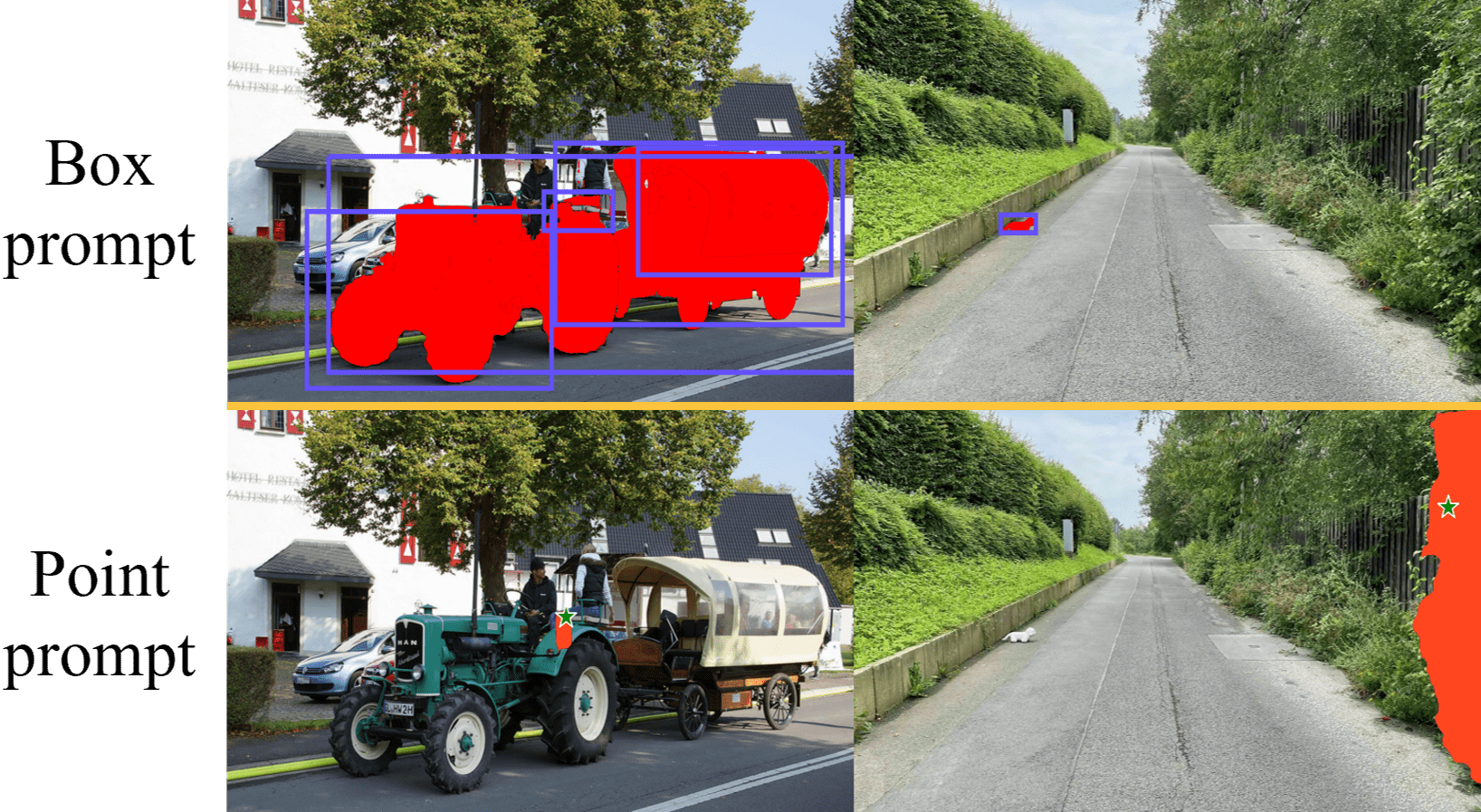}
    \caption{Box prompts can lead to \textbf{more accurate segmentation} for the OoD objects compared to point prompts. We derive point prompts from the locations corresponding to the extreme values of the anomaly scores. Visual analysis indicates that box prompts substantially improve the model's tolerance to noise. We employ all generated box prompts to obtain a comprehensive mask of the OoD object, which ensures that the entire object is covered.}
    \label{fig:boxpoint}
\vspace{-5pt}
\end{figure}

One possibility is to directly use pixels with high anomaly scores as prompts, which corresponds to \emph{point prompts}. However, point prompts, with their high level of specificity, carry a risk of significant inaccuracies. For example, due to their pinpoint nature, a point prompt may struggle to precisely locate the center of the OoD object which makes segmentation difficult. Even with multiple point prompts, achieving a good coverage of the OoD object remains a potential challenge. Fig. \ref{fig:boxpoint} shows that using point prompts for segmenting the OoD objects can lead to inaccurate masks.

Thus, we suggest creating box prompts that exhibit greater resilience to noise originating from anomaly scores. This approach enables more accurate segmentation of the OoD objects by the promptable segmentation model as shown in Fig. \ref{fig:boxpoint}. In particular, we leverage an object detector as the prompt generator to generate box prompt as the inputs for the promptable segmentation model. As the real-world OoD dataset is unavailable, we propose to train the prompt generator using dataset generated by outlier exposure.

\subsection{Outlier Exposure}
Outlier exposure (OE) has been widely used in OoD detection in semantic segmentation. For example, RPL \cite{RPL} leverages a synthetic training  dataset for training an OoD detector. Following this line of work, we propose to synthesize an OoD dataset for training the prompt generator using existing datasets.

In particular, suppose we have an inlier dataset as $\mathcal{D} ^{in} =  \left \{ \left ( x_{i}^{in} , y_{i}^{in}   \right )  \right \} _{i}^{\left |\mathcal{D}^{in}\right|}$, where $x^{in} \in \mathcal{X} \subset \mathbb{R} ^{H\times W\times3} $ represents the input image and $y^{in} \in \mathcal{Y}^{in}\subset\left \{0,1 \right \}^{H\times W} $ is the segmentation map with $C$ in-distribution categories. Similarly, the outlier dataset is defined as $\mathcal{D}_{out} = \left \{ \left ( x^{out}_{i},y^{out}_{i} \right )  \right \} _{i}^{\left |\mathcal{D}^{out}\right|} $, where $x^{out}\in\mathcal{X}$ and $y^{out}\in \mathcal{Y}^{out} \subset \left \{ 0, 1 \right \}^{H\times W}$ denotes the pixel-level mask label, with the class $1$ reserved for pixels belonging to the anomaly class. It should be noted that the $\mathcal{D}^{in}$and $\mathcal{D}^{out}$ do not have overlapping categories.

In the OE process, a transformation function $T$ is applied to the OoD object masks $y^{out}$. %This function adjusts the size and position of the OoD objects masks to align with the $\mathcal{D}^{in}$ image dimensions. 
It randomly rescales the OoD images and the corresponding masks at first. Then, it cuts or pads the images and masks into the dimension of the in-distribution images. The OE process is mathematically formulated as follows:
\begin{align}
x^{oe} = \left ( 1-T\left ( y^{out}\right )  \right ) \odot x^{in} +  T\left ( y^{out}\right )\odot x^{out}
\end{align}
Similarly, we also employ the transformation function $T$ to transform the outlier mask label,
\begin{align}
y^{oe} = T\left ( y^{out}\right )
\end{align}
%The input of the prompt generator is the anomaly score produced by the segmentation network on $x^{oe}$ and the output is the transformed outlier mask label $y^{oe}$.
The OE process will generate a synthetic OoD dataset $\mathcal{D} ^{oe} =  \left \{ \left ( x_{i}^{oe} , y_{i}^{oe}   \right )  \right \} _{i}^{\left |\mathcal{D}^{oe}\right|}$ for training the prompt generator.
%By this way, we can generate training dataset.
% reference or prompt result

\subsection{Prompt Generator Training }
The initial step of training the prompt generator involves transforming the mask labels $y^{oe}$ into prompt labels. In particular, a transformation function $T_{box}$ is used to generate box prompt $B_{prompt}$ from $y^{oe}$ by finding the coordinates of the smallest rectangular region that can fully contain all the masked pixels.
\begin{align}
B_{prompt}^{y^{oe}} = T_{box}\left (  y^{oe}\right ) 
\end{align}
Next, to obtain the anomaly scores for our synthetic training images, we utilize a mainstream OoD detection method $f_{ood}$. For each synthetic image $x^{oe}$, the anomaly score can be computed as,
\begin{align}
    S_{anomaly}^{x^{oe}} = f_{ood}\left ( x^{oe} \right ) 
\end{align}
With these anomaly scores and the bounding box prompts, we proceed to train our prompt generator, represented as $G_{prompt}$. The training objective is to enable  $G_{prompt}$ to process anomaly scores and generate corresponding bounding box prompts. The loss function can be written as:
\begin{align}
\mathcal{L}(\mathcal{D}^{oe}, G_{\text{prompt}}) = \sum_{x^{oe} \in \mathcal{D}^{oe}} \ell \left( G_{\text{prompt}}(S_{\text{anomaly}}^{x^{oe}}), B_{\text{prompt}}^{y^{oe}} \right)
\end{align}
Where $\ell$ is the loss function which quantifies the difference between the generated prompts and the actual box prompts. By minimizing the loss function $\mathcal{L}$, the prompt generator can generate accurate box prompts given the anomaly score map. The complete process is shown in Fig. \ref{fig:train}

In our experiments, we further enhance the robustness of the model by augmenting the anomaly scores with random noises. In particular, we achieve the best performance when the values of the anomaly score were subjected to a random fluctuation of $ 1\%$. This augmentation strategy not only improves the model's robustness to irregularities in the anomaly scores but also significantly boosted its generalization performance on real OoD datasets.

\begin{figure}[t]
    \centering
    \includegraphics[width=1.0\linewidth]{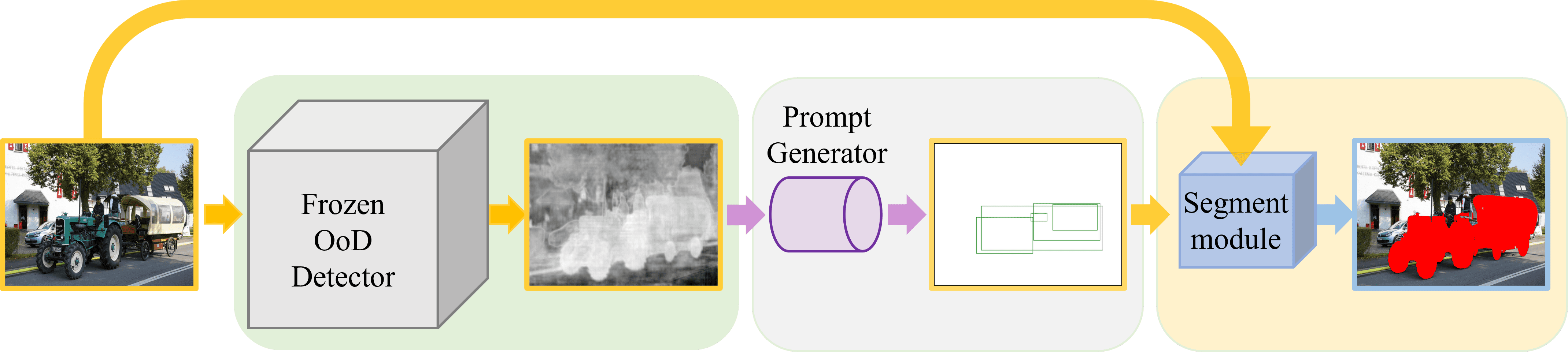}
    \caption{Overview of the inference pipeline.}
    \label{inference}
\vspace{-5pt}
\end{figure}
\subsection{Inference Pipeline}
During inference, the input image is processed using a state-of-the-art OoD detection method, to compute an anomaly score map. This map is then fed into the prompt generator, which yields bounding box prompts indicating the locations of potential OoD objects. Subsequently, we employ a promptable segmentation model which takes both the prompts and the original image as inputs and produces masks of the OoD objects as shown in Fig. \ref{inference}.

Since the regions with high anomaly scores can be fragmented, it is possible that the prompt generator would produce multiple box prompts. In this case, we feed all of them into the promptable segmentation model and use the union area of all generated masks as the final result.

\section{Experiments}
\begin{table*}[!ht]
\small
\setlength{\tabcolsep}{1.5pt}
\begin{tabular*}{\linewidth}{@{\extracolsep{\fill}}r| ccc | ccc | ccc | ccc | cccc@{}}
\toprule
\multicolumn{1}{c}{\multirow{2}{*}{\makecell{Methods}} }
& \multicolumn{3}{c}{FS Static} & \multicolumn{3}{c}{FS Lost\&Found} & \multicolumn{3}{c}{SMIYC-Anomaly} & \multicolumn{3}{c}{SMIYC-Obstacle} & \multicolumn{3}{c}{RoadAnomaly} \\
\cmidrule(lr){2-4} \cmidrule(lr){5-7} \cmidrule(lr){8-10} \cmidrule(lr){11-13} \cmidrule(lr){14-16}
& \makecell{AuIoU} & \makecell{IoU} & \makecell{mean F1} & \makecell{AuIoU} & \makecell{IoU} & \makecell{mean F1} & \makecell{AuIoU} & \makecell{IoU} & \makecell{mean F1} & \makecell{AuIoU} & \makecell{IoU} & \makecell{mean F1} & \makecell{AuIoU} & \makecell{IoU} & \makecell{mean F1} \\
\midrule
DenseHybrid \cite{Densehybrid}& 6.81 & 23.54 & 11.15 & 2.41 & 15.35 & 3.83 & 19.25 & 35.77 & 29.4 & 2.23 & 20.04 & 3.41 & 13.53 & 26.17 & 21.34 \\
Synboost \cite{Synboost}& 17.89 & 32.81 & 25.67 & 7.37 & 18.39 & 10.85 & 28.35 & 42.00 & 39.48 & 5.45 & 13.97 & 9.02 & 19.50 & 27.22 & 29.33 \\
PEBAL \cite{PEBAL} & 8.21 & 26.92 & 13.31 & 1.49 & 6.35 & 2.57 & 23.98 & 42.39 & 35.09 & 0.57 & 6.74 & 1.06 & 15.51 & 33.80 & 23.87 \\ 
RPL+CoroCL \cite{RPL}& 8.46 & 36.46 & 13.16 & 2.55 & 15.75 & 3.91 & 22.26 & 68.80 & 31.55 & 3.85 & 28.66 & 5.72 & 16.88 & 50.97 & 24.64 \\
%S2M (Ours) & \textbf{63.98} & \textbf{70.77} & \textbf{70.51} & \textbf{30.97} & \textbf{32.24} & \textbf{37.21} & \textbf{58.12} & \textbf{83.48} & \textbf{64.86} & \textbf{51.26} & \textbf{60.63} & \textbf{56.53} & \textbf{49.04} & \textbf{58.73} & \textbf{55.67} \\ %2% noise
S2M (Ours) & \textbf{64.78} & \textbf{69.99} & \textbf{70.24} & \textbf{29.30} & \textbf{30.46} & \textbf{35.31} & \textbf{54.33} & \textbf{77.54} & \textbf{60.40} & \textbf{58.12} & \textbf{67.64} & \textbf{64.96} & \textbf{54.31} & \textbf{58.49} & \textbf{61.66} \\ %1% noise
\bottomrule
\end{tabular*}
\caption{\textbf{S2M outperforms the state-of-the-art methods on Fishyscapes Static (FS Static), Fishyscapes Lost and Found (FS Lost\&Found), SMIYC Anomaly, SMIYC Obstacle and RoadAnomaly validation set}. In our method, which is based on the anomaly score from RPL+CoroCL, the IoU is calculated without a threshold and by using all the produced prompt boxes. Both AuIoU and mean F1 scores are calculated by iterating through all thresholds. To facilitate a fair comparison with existing methods, we use the confidence of the box prompt as the anomaly score when calculating AuIoU and mean F1.}
\label{table1_compareSOTA}
\vspace{-10pt}
\end{table*}
\subsection{Experimental Setup}

\noindent\textbf{Datasets.} We evaluate our method on several OoD detection benchmarks. Fishyscapes \cite{L&F} is an urban driving scenes dataset consists of two datasets, Fishyscapes static (FS static) and Fishyscapes lost \& found (FS lost \& found), which has high-resolution images for anomaly detection. Segment-Me-If-You-Can (SMIYC) \cite{segmentmeifyoucan} benchmark comprises two distinct datasets: \emph{RoadAnomaly} (RA) and \emph{RoadObstacle} (RO), designed for evaluating the performance of models in segmenting road anomalies and obstacles. SMIYC-RA features 110 images with diverse anomalies, while SMIYC-RO contains 442 images, focusing on small objects on roads, including challenging conditions like nighttime and adverse weather. %Both datasets present a significant domain shift compared to Cityscapes, adding to their complexity. 
The Road Anomaly \cite{RoadAnomaly} dataset, precursor to SMIYC, features 60 diverse images for real-world anomaly detection, including a validation set with internet-sourced anomalies.

\noindent\textbf{Outlier Exposure.} We follow the previous work \cite{OE, chan2021entropy, RPL} for outlier exposure (OE) and leverage Cityscapes \cite{cityscapes}  as the inlier dataset and COCO \cite{COCO} as the outlier dataset to generate synthetic training images. Cityscapes consists of 2975 images for training. There are a total of 19 classes which are viewed as inlier categories. Objects in COCO dataset are used as OoD objects. For a fair comparison, we generate the same number OE images as in \cite{RPL} for training the prompt generator.
%And we generate the prompt label from mask label as method mentioned above. In order to make our experiment easier to recurrent, we use the fixed offline dataset instead of online generating dataset, which is difference each time when it process outlier exposure.

\noindent\textbf{Baselines.} We compared our method with the OoD detection methods, 
\begin{itemize}
    \item \textbf{RPL} \cite{RPL}. It proposes a residual pattern learning module and employs a context-robust contrastive learning method to enhance the capability of OoD detection.
    \item  \textbf{PEBAL} \cite{PEBAL}. This method introduces pixel-wise energy-biased abstention learning, which synergistically optimizes a novel pixel-wise anomaly abstention learning framework along with energy-based models.
    \item \textbf{Synboost} \cite{Synboost}. This framework enhances re-synthesis methods using uncertainty maps to identify mismatches between generated images and their original counterparts. 
    \item \textbf{DenseHybrid} \cite{Densehybrid}. Densehybrid is implemented by integrating generative modeling of regular training data with discriminative analysis of negative training data, aiming to create a hybrid algorithm that balances the strengths and addresses the weaknesses of both approaches.
\end{itemize}
\noindent\textbf{Evaluation Metrics.}
We leverage component-level metrics including Intersection over Union (IoU) \cite{GIOU} and mean F1 \cite{segmentmeifyoucan}, along with pixel-level metrics including area under the precision-recall curve (AuPRC) \cite{AUROC_Baur} and false positive rate at a true positive rate of 95\% (FPR95) \cite{Synboost}, to compare various methods. Specifically, component-level metrics assess the quality of OoD object masks, while pixel-level metrics focus on the effectiveness of anomaly detection for individual pixels.

Moreover, as the quality of the masks generated by anomaly score-based detection methods relies heavily on the threshold. We further introduce a new component-level metric called Area under IoU Curve (AuIoU) for evaluating the sensitivity of different methods to the selection of threshold. 
We calculate AuIoU by computing the IoU across all thresholds ranging from 0 to 1 in increments of 0.01 and then determining the area under the IoU curve. In particular, a higher AuIoU value for a method indicates not only excellent IoU performance but also the ease of selecting an appropriate threshold.

%To compare the difficulty of selecting thresholds by different methods, we designed an Area under IoU Curve(AuIou). A larger area under the IoU curve indicates a relative ease in finding a suitable threshold. Additionally, we employ the mean F1 score, calculated as an average across all thresholds, to evaluate the overall performance of our method, as referenced in \cite{segmentmeifyoucan}. We also evaluate the area under precision recall curve (AuPRC) and false positive rate at a true positive rate of 95\% (FPR95) on SMIYC\cite{segmentmeifyoucan}.

\noindent\textbf{Implementation Details.} Our implementation is derived from \cite{detectron2,RPL}. We use a faster R-CNN \cite{fasterrcnn} as a prompt generator with ResNet-50 \cite{resnet} as the backbone. The ResNet-50 backbone is pre-trained on ImageNet \cite{imagenet}. We leverage the anomaly scores generated by RPL \cite{RPL} for training and inference, which employs DeepLabv3+ as segmentation model with WiderResNet38 backbone. 

During the training process, we initiate the learning rate at $1\times 10^{-4}$, employing a learning rate adjustment strategy. This approach incrementally raises the learning rate in a linear fashion to reach $2.5\times 10^{-3}$ over the course of 1000 iterations. We train our prompt detector on 1 $\times$ NVIDIA RTX A5000 GPU within about 10 hours for 100 epochs.
%To facilitate the reproducibility of our experiments, we trained our model using a fixed dataset of 2966 preserved images, rather than employing random generation at each iteration as previous methods have done. 
We employ the Segment Anything Model (SAM) \cite{SAM} as our promptable segmentation model, utilizing a ViT\-B backbone. SAM processes both the original image and box prompts from the prompt generator for accurately segmenting the OoD objects.

To enable other methods to achieve their best performance, we identify the optimal threshold on the validation dataset for each dataset. We search for the optimal threshold  over a range of thresholds by varying $t$ from 0 to 1 with 0.01 as the step size as follows,
\begin{align}
    t_{real} = t\times (S_{max}-S_{min}) + S_{min}
\end{align}
where $S_{max}$ and $S_{min}$ are the minimum and the maximum anomaly scores on the corresponding dataset, respectively. The $t^*_{real}$ which achieves the best IoU is used for the final threshold for generating the masks of the OoD objects. In particular, for each pixel $i$ if $S_i(x; \theta)>t^*_{real}$, then the pixel is viewed as part of the OoD object. For the baselines, we report the IoU achieved by the optimal thresholds.

For our method, we compute IoU without relying on a threshold and utilizing all masks generated from the produced box prompts. However, computing the mean F1 and AuIoU requires a varying threshold. In this case, we leverage the confidence score generated by the Faster R-CNN for each box prompt. We assign the confidence score of each box prompt as the anomaly scores of the pixels within the corresponding mask. For pixels within multiple overlapping masks, we adopt the lowest confidence score across all the masks to reduce the false positive rate when using a small threshold.

\subsection{Main Results} 
\textbf{Component-level metrics.} Table \ref{table1_compareSOTA} shows the results of S2M and the baselines on the benchmark datasets. The results show that the proposed S2M achieve superior results in terms of IoU, AuIoU and mean F1 score, indicating that S2M can accurately segment the OoD objects.

The results also show that existing anomaly score-based OoD detection methods, such as DenseHybrid, RPL, and PEBAL, are less effective in producing accurate masks for the OoD objects. While the RPL+CoroCL method achieves high IoU values, it does not perform as well across other metrics, such as the mean F1 score \cite{segmentmeifyoucan}, revealing a shortfall in achieving highly accurate masks for OoD objects \cite{segmentmeifyoucan}. Our method, however, excels in all assessed metrics. The IoU achieved by our method surpasses that of the state-of-the-art methods by approximately 20\% across all datasets, on average. Specifically, on the SMIYC-Obstacle validation dataset, our method outperforms RPL by a significant margin of 38.98\%.

%As illustrated in Figure \ref{fig:MainCompare}, our approach delivers the most accurate mask results, surpassing those of RPL and PEBAL, and aligning with experimental results. During visualizations, we set the thresholds for other methods to their optimal IoU performance on each dataset. However, a limitation becomes evident in images like the sloth picture, which is not recognized accurately due to a relatively low threshold for that specific image. This highlights a persistent issue of distribution shift within a single dataset, impacting the reliability of anomaly scores in real-world scenarios.

\begin{figure}[tbp]
	\centering
        \scriptsize
	\begin{subfigure}{0.495\linewidth}
		\centering
		\includegraphics[width=1\linewidth]{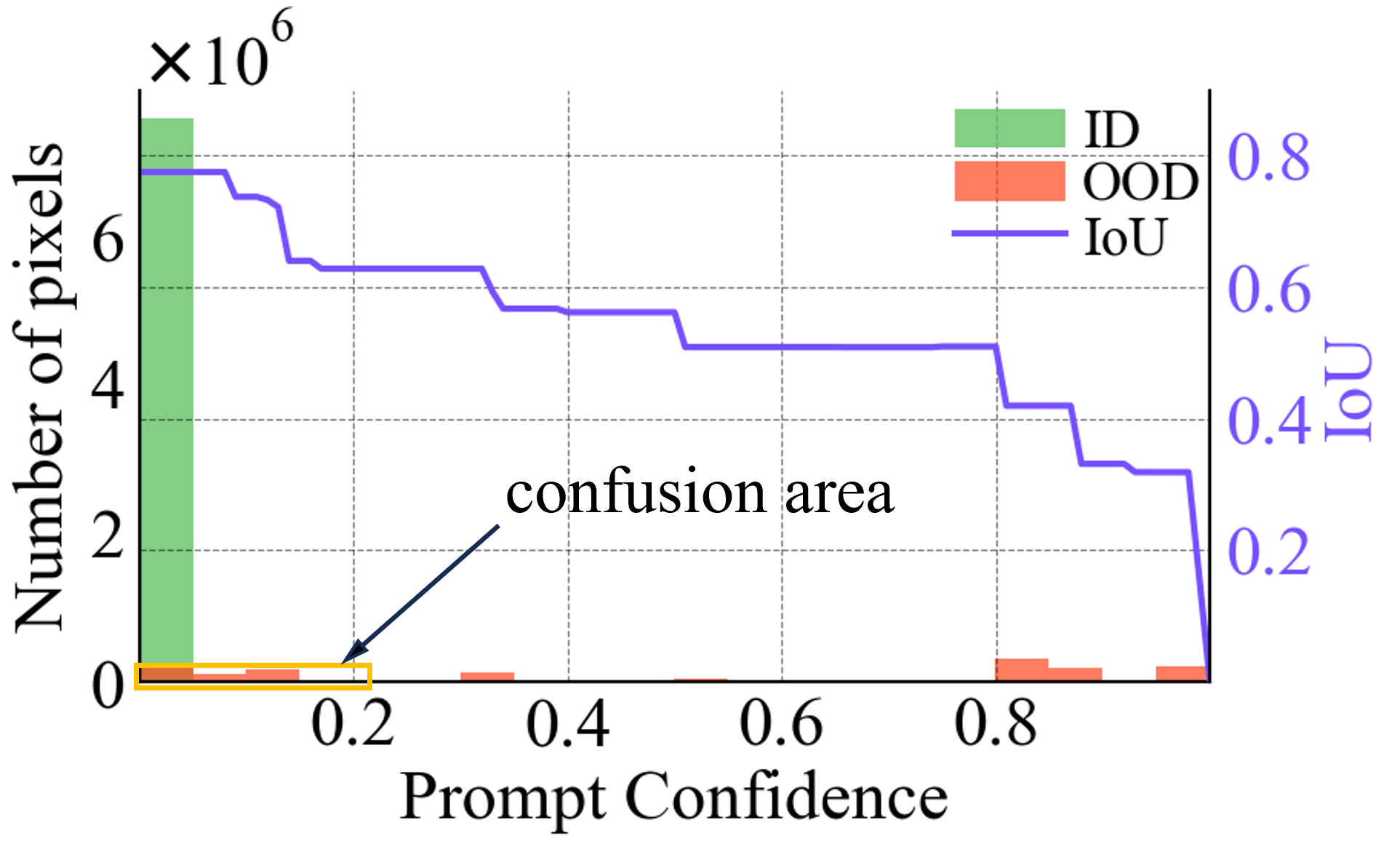}
		\caption{S2M}
		\label{test1}
	\end{subfigure}
	\centering
	\begin{subfigure}{0.495\linewidth}
		\centering
		\includegraphics[width=1\linewidth]{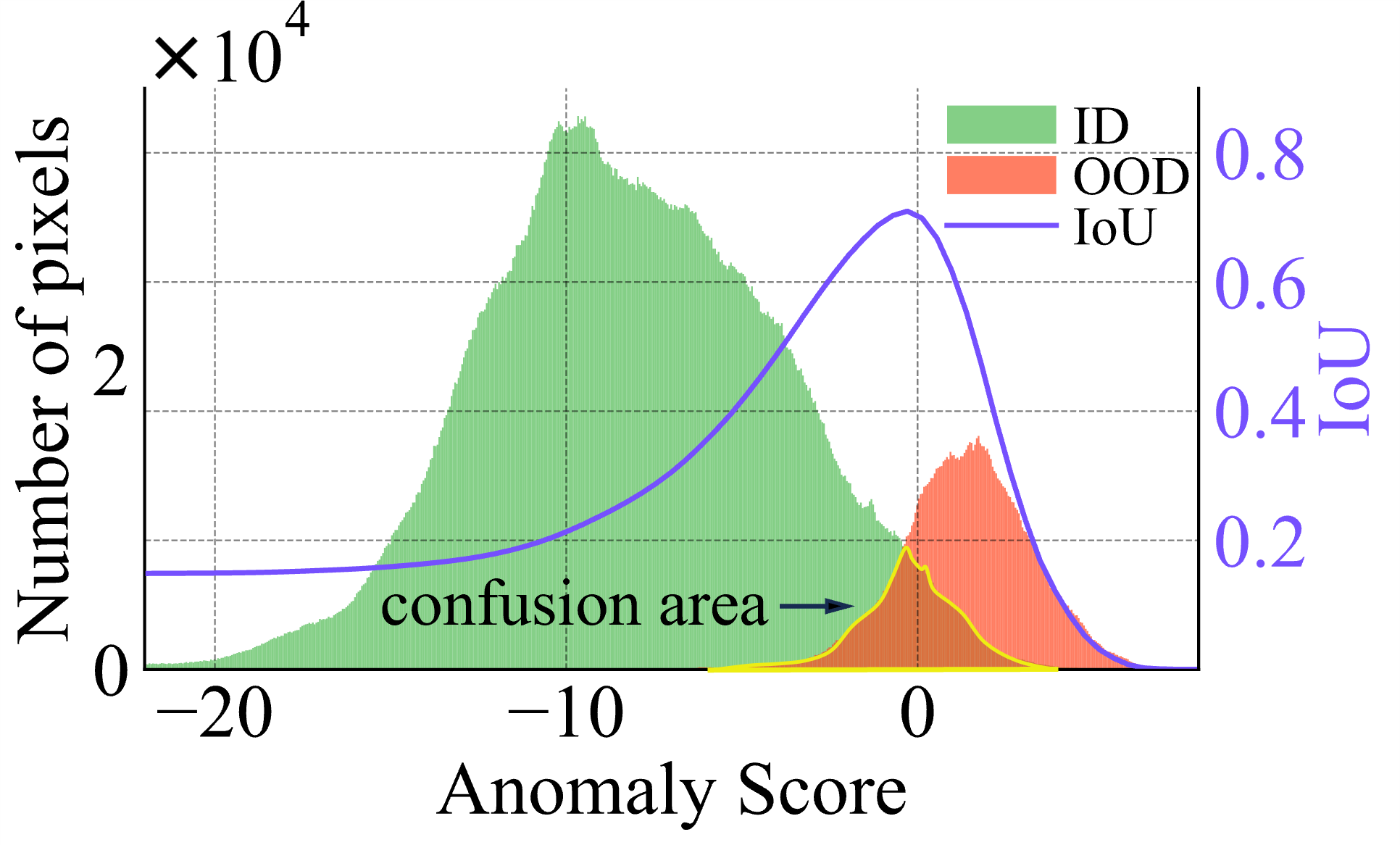}
		\caption{RPL}
		\label{test2}
	\end{subfigure}
	\caption{\textbf{Anomaly score distribution and IoU curve} of our method and RPL on SMIYC RoadAnomaly validation dataset reveals that S2M excels in identifying both ID and OoD pixels, with a diminished area of confusion relative to RPL. Contrasting with traditional approaches, S2M bypasses threshold selection and directly achieves optimal IoU with all generated box prompts.}
	\label{distributon}
\vspace{-10pt}
\end{figure}

\begin{table}[]
    \centering
    \scriptsize
    \begin{tabular}{@{}ccccccc@{}}
    \toprule
    {}  & \multicolumn{2}{c}{{SMIYC anomaly}}  & \multicolumn{2}{c}{{SMIYC obstacle}}  & \multicolumn{2}{c}{{RoadAnomaly}} \\ \cmidrule(l){2-3} \cmidrule(l){4-5}  \cmidrule(l){6-7}
    {Noise} & {IoU}   & {mean F1}       & {IoU}    & mean F1  & {IoU}   & mean F1   \\ \cmidrule(r){1-7}
    {0\%}& {74.57}& {63.15}& {66.29} & {62.61} & {55.68} & {52.89}\\
    {1\%} &{77.54}&{60.40}&{\textbf{67.64}}&\textbf{64.96}&58.49&\textbf{61.66}\\
    {2\%} & {\textbf{83.44}} & {\textbf{65.60}} & {60.62}  &56.41  & {\textbf{58.67}} & 55.46\\
    {3\%} & {74.89}& {59.92}& {56.53} & {48.11} & {56.18} & {54.93}\\
    {4\%} & {77.85}& {54.59}& {57.73} & {57.67} & {56.72} & {56.45}\\
    %{5\%} & {73.02} & {\textbf{70.81}}& {66.42} & {71.64} & {57.05} & {57.15}\\ \bottomrule 67.13 67.115
    \bottomrule
    \end{tabular}
    \caption{Performance across different \textbf{noise intensities.} We introduced a random perturbation to the magnitude of the anomaly scores and observed that a fluctuation of 1\% yielded the best training outcomes on these datasets. All experimental results reflect the performance after 80 epochs of training. The mean F1 scores are derived by systematically iterating through all thresholds in increments of 0.01.}
    \label{Noise}
\vspace{-10pt}
\end{table}

\textbf{Pixel-wise metrics.} To thoroughly evaluate the efficacy of our proposed method, we also evaluate the AuPRC and FPR95 on the SMIYC validation set. We leverage the confidence score from the prompt generator to calculate both AuPRC and FPR95. For those area covered by more than one mask, we use the lowest score as discuss before. As evidenced by Table \ref{table2_compareSMIYC}, our approach yields notable results, maintaining strong performance across detailed pixel-level metrics. Notably, the FPR95 of our approach is much lower compared to other methods, aligning with the expected reduction of false positive results. Further examination of Fig. \ref{distributon} reveals that the area of confusion in our method's results is substantially reduced relative to that of RPL \cite{RPL}. This underscores our method's enhanced capability in discriminating between in-distribution (ID) pixels and OoD pixels.

\begin{table}[!ht] 
\scriptsize
\setlength{\tabcolsep}{2pt}
\begin{tabular*}{\linewidth}{@{\extracolsep{\fill}}rcccc@{}}
\toprule
\multicolumn{1}{c}{\multirow{2}{*}{\makecell{Methods}}} & \multicolumn{2}{c}{SMIYC Anomaly} & \multicolumn{2}{c}{SMIYC Obstacle} \\
\cmidrule(lr){2-3} \cmidrule(lr){4-5}
 & \makecell{AuPRC} & \makecell{FPR↓} & \makecell{AuPRC} & \makecell{FPR↓} \\
\midrule
Maximum softmax \cite{Dan_Baseline}& 40.4 & 60.20 & 43.4 & 3.80 \\ 
Mahalanobis \cite{Mahalanobis}& 22.50 & 86.40 & 25.9 & 26.10 \\ 
SML \cite{standardizedMaxLogits} & 21.68 & 84.13 & 18.60 & 91.31 \\ 
Synboost \cite{Synboost}& 68.80 & 30.90 & 81.40 & 2.80 \\ 
Meta-OoD \cite{chan2021entropy}& 80.13 & 17.43 & 94.14 & 0.41 \\ 
DenseHybrid \cite{Densehybrid}& 61.08 & 52.65 & 89.49 & 0.71 \\ 
PEBAL \cite{PEBAL}& 53.10 & 36.74 & 10.45 & 7.92 \\
RPL+CoroCL \cite{RPL}& 88.55 & 7.18 & \textbf{96.91} & 0.09 \\ 
\midrule
S2M (Ours) & \textbf{91.92} & \textbf{1.04} & 91.73 & \textbf{0.02} \\ 
\bottomrule
\end{tabular*}
\caption{\textbf{Pixel-level metrics.} Our S2M method, evaluated against state-of-the-art approaches using pixel-level metrics including FPR95, shows comparable OoD detection capabilities with a significantly lower FPR.}
\label{table2_compareSMIYC}
\vspace{-10pt}
\end{table}

\subsection{Ablation Studies}

\begin{figure}[h!tbp]
	\centering
	\begin{subfigure}{0.495\linewidth}
		\centering
		\includegraphics[width=1\linewidth]{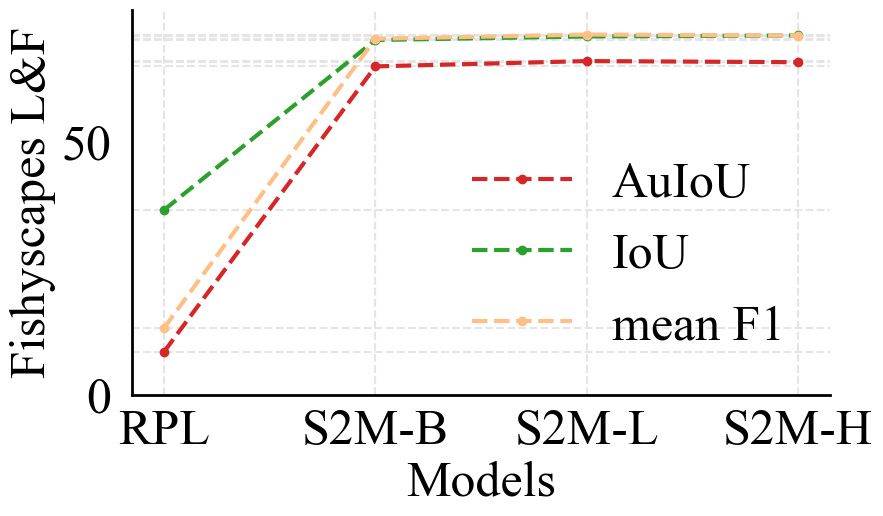}
		\caption{Fishyscapes Static}
		\label{Static}
	\end{subfigure}
	\centering
	\begin{subfigure}{0.495\linewidth}
		\centering
		\includegraphics[width=1\linewidth]{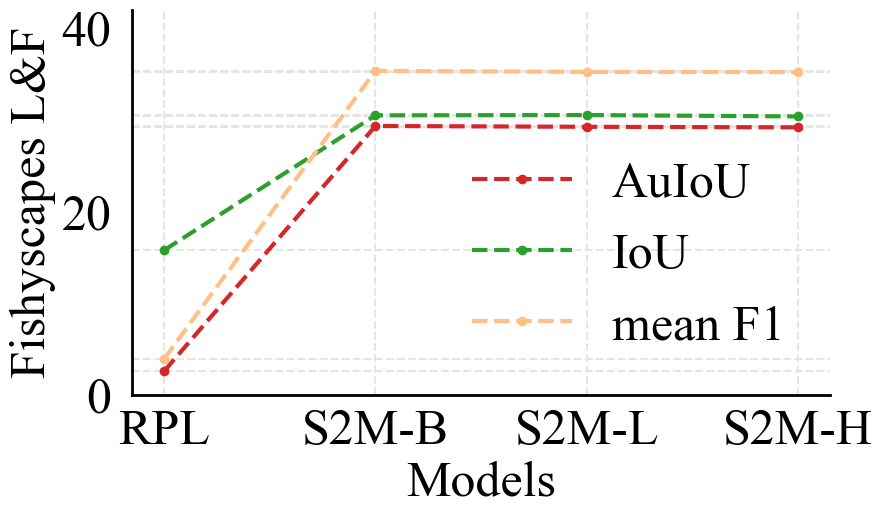}
		\caption{Fishyscapes L\&F}
		\label{scale of model}
	\end{subfigure}
	\caption{\textbf{S2M is robust to the different size of SAM.} The results from various versions of S2M on Fishyscapes Static and Fishyscapes L\&F indicate that a smaller SAM is able to achieve high segmentation performance.}
	\label{scability}
\vspace{-10pt}
\end{figure}

\noindent\textbf{Generalize to other anomaly scores.}
To further validate the generalization of our method, we leverage PEBAL \cite{PEBAL} for generating the anomaly scores in S2M. Applying PEBAL to generate anomaly scores for the synthetic training dataset and utilizing the previously mentioned training strategy, we also observed similar performance improvement. Table \ref{pebals2m} shows that a significant increase in terms of IoU, with improvements of 12.12\% and 34.03\% on the SMIYC anomaly and obstacle tracks \cite{segmentmeifyoucan} compared with PEBAL , respectively. This shows that S2M can generalize to various anomaly scores and be easily integrated with existing anomaly score-based OoD detection methods.

\noindent\textbf{Data augmentation.}
We experimented with different levels of random fluctuations within a specific range to the values of the anomaly scores and subsequently tested the model's performance on the Fishyscapes Static and Fishyscapes Lost \& Found \cite{Fishyscapes} datasets. Table \ref{Noise} indicate that the model achieves optimal overall performance with the addition of 1\% noise. The model exhibited a 2.97\% increase in the best IoU on the SMIYC anomaly dataset, and there was also an improvement in the mean F1 scores across both datasets. On the SMIYC obstacle dataset, the model performance saw a slight improvement, with an increase of 1.35\% in the best IoU. Furthermore, it was observed that an excessive noise level leads to a significant reduction in the mean F1 score, suggesting that overly strong noise can adversely affect model training, hindering the model's ability to accurately generate prompts.

\noindent\textbf{Different choices of SAM.}
We investigated various SAM backbone configurations, including ViT-B, ViT-L, and ViT-H \cite{SAM}, the corresponding S2M is denoted as S2M-B, S2M-L and S2M-H as shown in Fig. \ref{scale of model}. When employing SAM with ViT-B as the backbone in S2M, the best IoU is improved by 33.53\% and 14.71\% on the Static and L\&F datasets than RPL, respectively. Comparatively, the IoU improvement from S2M-H to S2M-B is modest, with increases of 0.89\% and 0.03\% on Static and L\&F. This pattern suggests that S2M remains robust across various SAM choices, owing to its reliance on box prompts for mask generation. 
%that the observed performance gains are attributable more to the S2M pipeline itself rather than to the inherent strength of the ViT\_L or ViT\_H backbones. The substantial improvement in S2M-B Over RPL further demonstrates the efficacy of our approach.

\begin{figure}
    \centering
    \includegraphics[width=0.8\linewidth]{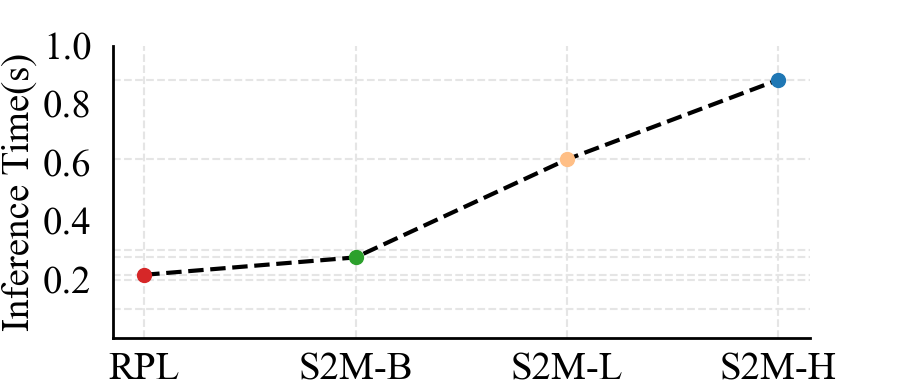}
    \caption{\textbf{Runtime analysis} for RPL and various scales of S2M base on RPL. The runtime of RPL is included in S2M. The findings reveal that S2M introduces only a minimal increase in processing time, underscoring its viability for practical deployment scenarios.}
    \label{runtime}
\vspace{-5pt}
\end{figure}

\begin{table}[]
    \centering
    \scriptsize
    \begin{tabular}{@{}cccc@{}}
    \toprule
        {Datasets}     & {Metrics}           & {PEBAL} & {S2M (PEBAL)} \\ \midrule
        {}     & {AuIoU}       & {23.98}  & \textbf{28.40}    \\
        {Anomaly} & {IoU} & {42.39} & \textbf{54.51}     \\
        {}     & {mean F1}     & {35.09} & \textbf{37.03}    \\ \midrule
        {}     & {AuIoU}       & {0.57}  & \textbf{24.27}    \\
        {Obstacle} & {IoU} & {6.74} & \textbf{40.77}    \\
        {}     & {mean F1}     & {1.06}  & \textbf{27.07}    \\ \midrule
        {}     & {AuIoU}       & {15.51} & \textbf{29.42}    \\
        {RoadAnomaly}   & {IoU} & {33.80}  & \textbf{36.17}    \\
        {}     & {mean F1}     & {23.87} & \textbf{33.30}    \\ \bottomrule
    \end{tabular}
    \caption{\textbf{S2M can generalize to other anomaly scores.} S2M enhances OoD segmentation performance by incorporating the anomaly score from PEBAL as input, demonstrating a significant improvement over PEBAL.}
    \label{pebals2m}
\vspace{-10pt}
\end{table}

\subsection{Efficiency Analysis}
For efficiency analysis, we first measured the standalone execution speed of RPL, followed by assessments of S2M-B, S2M-L, and S2M-H in Fig. \ref{runtime}. This evaluation involved processing all images in the Road Anomaly dataset and calculating the time taken for each image. The experiments were conducted on an NVIDIA RTX A5000 GPU. The runtime measurement encompassed the duration from the model receiving the input to producing the anomaly score or mask. For the RPL model, the processing time per image was recorded at 0.217s. Comparatively, S2M-B processed each image in 0.276s, showing only a marginal increase of 0.059s, which amounts to a 27.2\% longer duration. The parameter of RPL model is 168M, and the total parameter of our S2M, built upon RPL, amounts to 300M. These results indicate that our method does not significantly burden resources during inference, thereby demonstrating its strong potential for practical deployment.

\begin{figure}
    \centering
    \includegraphics[width=1\linewidth]{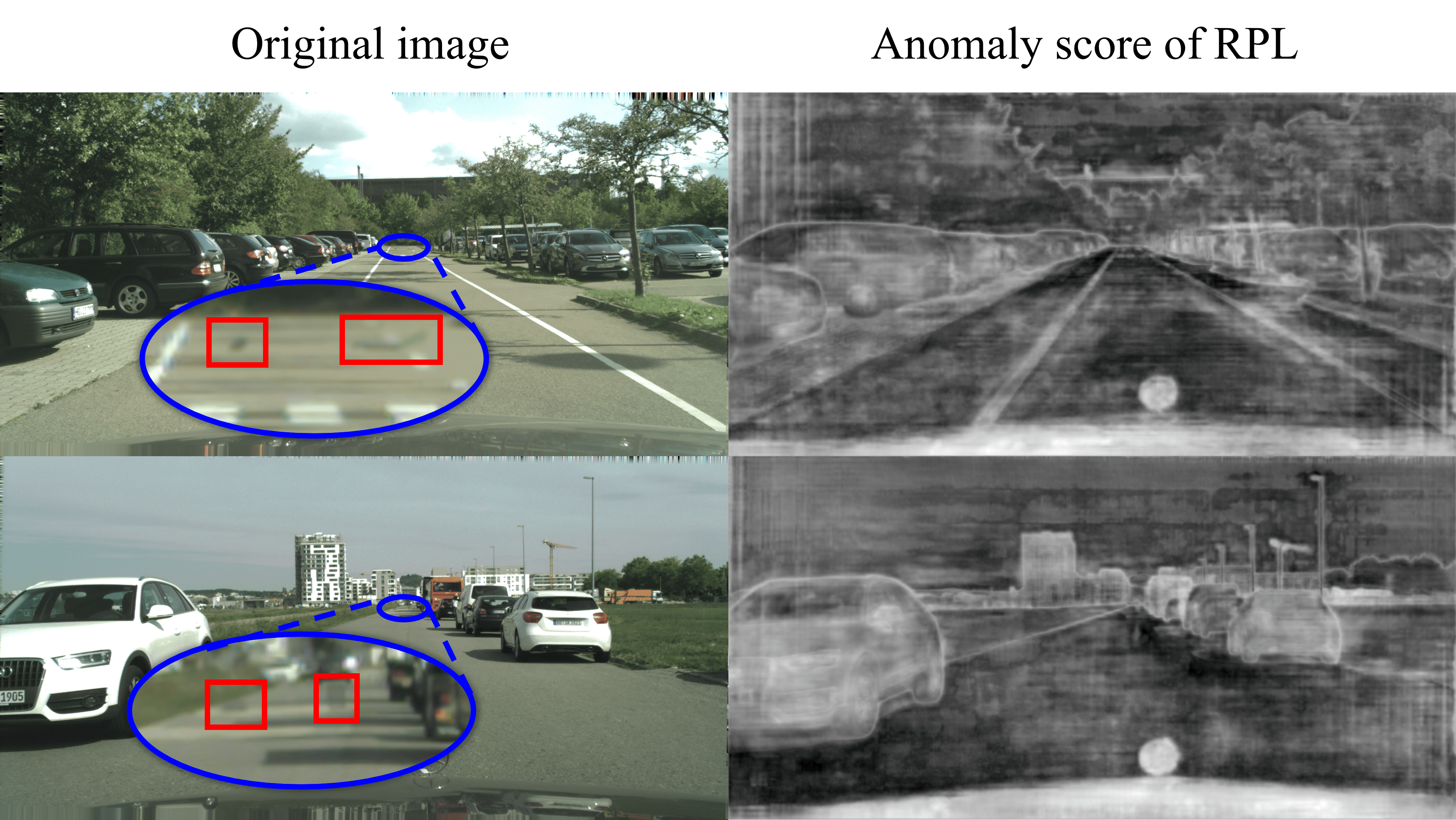}
    \caption{\textbf{Failure cases} for S2M. Challenges arise when OoD objects are too small for current detectors to identify, leading to anomaly score maps lacking distinctive features. The left column shows original images, highlighting OoD objects with a magnifying glass, the right column showing the anomaly scores maps.}
    \label{failure}
\vspace{-10pt}
\end{figure}
\subsection{Failure cases}
Although our method demonstrates effective performance in most scenarios, there are still a few instances where it fails during inference. These failures occur when the prompt generator is unable to generate box prompts from the anomaly score map. We have visualized some of these failure cases for a better understanding. In Fig. \ref{failure}, the right column displays masks corresponding to the best IoU values. Notably, the best IoU for RPL's anomaly scores on these two images are only 0.03\% and 0.01\%, respectively. This low detection rate is attributed to the OoD objects being extremely small and translucent in these images, making them challenging to recognize. Consequently, S2M may fail to recognize any anomalies when the anomaly score map does not contain meaningful information for indicating the location of the OoD objects. For future work, we aim to investigate more robust training strategies to effectively mitigate inaccuracies in anomaly scores.

%%%%%%%%%%%5%% 子图

% \begin{table}[]
%     \centering
%     \scriptsize
%     \setlength{\tabcolsep}{1.5pt}
%     \begin{tabular}{@{}cccccccccc@{}}
%     %\toprule
%     {}  & \multicolumn{3}{c}{{SMIYC anomaly}}  & \multicolumn{3}{c}{{SMIYC obstacle}}  & \multicolumn{3}{c}{{RoadAnomaly}} \\ %\cmidrule(l){2-4} \cmidrule(l){5-7}  \cmidrule(l){8-10}
%     {Noise} & {IoU}   & {meanF1}  & {AuIoU}   & {IoU}   & {meanF1}  & {AuIoU} & {IoU}   & {meanF1}  & {AuIoU} \\ \cmidrule(r){1-10}
%     {PEBAL}& {77.23}& {61.74}& {66.02} & 67.02 & 55.15 & 52.13\\
%     {S2M(PEBAL)} &{79.20}&{61.06}&{\textbf{71.70}}&\textbf{72.42}&\textbf{59.18}&\textbf{59.13}\\ \bottomrule
%     \end{tabular}
%     \caption{Performance across different \textbf{noise intensities.} we introduced a random perturbation to the magnitude of the anomaly scores and observed that a fluctuation of 1\% yielded the best training outcomes. All experimental results reflect the performance after 80 epochs of training. The mean F1 scores are derived by systematically iterating through all thresholds in increments of 0.01.}
%     \label{Noise}
% \end{table}

\section{Summary}
We introduce S2M, a simple and effective pipeline for OoD detection in semantic segmentation. S2M converts any anomaly score map into segmentation masks for accurately segmenting the OoD objects. S2M is general, capable of integrating anomaly scores from various OoD detectors. Extensive experiments demonstrate that our method surpasses other state-of-the-art OoD detection techniques on several commonly used OoD detection benchmarks datasets in terms of both component-level metrics as well as pixel-level metrics. %We believe that S2M can be easily integrated in existing autonomous systems for accurately detecting OoD objects, thereby contributing to the overall robustness of these systems.

{\bf Acknowledgements.} 
We would like to thank the anonymous reviewers for their helpful comments. This project was supported by a grant from the University of Texas at Dallas.

{
    \small
    \bibliographystyle{ieeenat_fullname}
    \bibliography{main}
}


\clearpage
\setcounter{page}{1}
\maketitlesupplementary

\section{Supplementary}
\label{sec:supp}

We propose the first prompt-based OoD detection method.  Our core idea has two main aspects: {\bf 1)} Generating prompts directed at OoD objects using information from the anomaly score map and {\bf 2)} employing a prompt-based segmentation model to provide accurate masks for OoD objects. In this phase, the segmentation model, refined through prompts, accurately identifies and segments out OoD objects, enhancing the overall detection accuracy and efficiency. Together, these novel steps demonstrate exceptional performance in the field of OoD detection, offering a new perspective for the identification of OoD objects. In this supplementary, we include more details on the following aspects:
 
\begin{itemize}
    \item We present the implementation details of acquiring training data using the OE method in Section~\ref{subsec:OE}.
    \item We delineate the specifics of generating OoD object masks in Section~\ref{subsec:MG}.
    \item We provide a detailed description of the primary evaluation metrics used in our experiments, elucidating the significance of each metric and the performance of our S2M method across these metrics in 
 Section~\ref{subsec:EM}.
 
    \item We detail the efficiency analysis, demonstrating the operational effectiveness of our approach in Section~\ref{subsec:EA}.
 
    \item We employ FastSAM in stead of the standard SAM in Section~\ref{subsec:FastSAM}.
    \item We utilize an entropy-based anomaly score in Section~\ref{subsec:EB}.
    \item We present visualizations of some S2M results in Section~\ref{subsec:Visualization}.
\end{itemize}

\subsection{Details of Outlier Exposure}
\label{subsec:OE}
\begin{figure*}[p]
    \centering
    \includegraphics[width=0.8\textwidth]{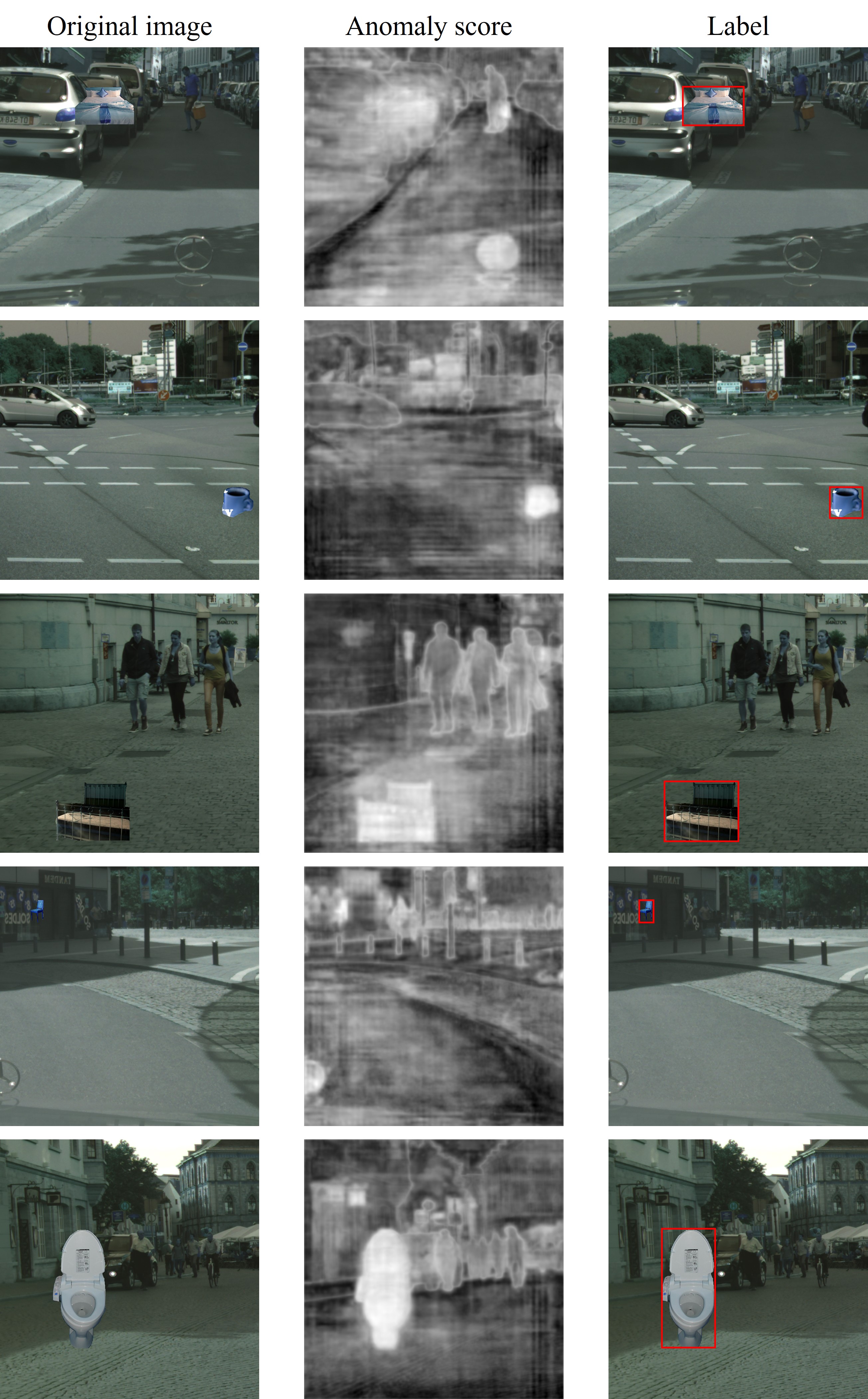}
    \caption{Visualization of training data.}
    \label{contact}
\end{figure*}

During the preparing of training dataset, we use the OE \cite{OE,RPL} strategy to generate the OoD training images. We use objects in COCO dataset as OoD objects, and use images in Cityscapes dataset as background. We exclude those objects from the COCO that are also included in the Cityscapes. The left column of Fig. \ref{contact} shows the generated training image. Then, we use RPL \cite{RPL} to get the anomaly score on these training images. The anomaly scores of these training images are shown in the middle column. The original anomaly scores, which generally lie between -20 and 10, are not suitable for visualization. For visualization purposes, we have normalized these scores to a scale of 0 to 255 for each image. It should be noted that the training process uses the original anomaly scores, not the normalized ones. Right column show the training label of OoD objects. We generated the smallest bounding boxes based on the masks of the OoD objects, which serve as the training labels. During the training of the prompt generator, we utilize the anomaly scores as inputs and employ the generated boxes as prompts.

\subsection{Details of Mask Generation}
\label{subsec:MG}
During the inference, we use the produced box prompts to generate masks of OoD objects. The prompt generator is designed to process anomaly scores as input, thereby generating box prompts that highlight OoD objects. In addition, it concurrently produces confidence scores associated with these prompts. To enable a direct comparison between our S2M method and current mainstream approaches using the same metrics, the corresponding confidence scores of these prompts are assigned to the pixels in the generated masks for the OoD objects. For areas with multiple overlapping masks, the pixel values are assigned based on the lowest confidence score among the box prompts that produced these overlapping masks. We employ this strategy with the intention of lowering the false positive rate. Ultimately, the output of our S2M methods is a map where pixel values ranging from 0 to 1. In this map, a pixel value of 0 indicates ID areas, while any other values correspond to OoD regions.

\subsection{Evaluation Metrics}
\label{subsec:EM}
During the experimental process, we employed three evaluation metrics. The first metric, IoU, is used to assess the accuracy of OoD object detection at a specific threshold. However, since IoU does not reflect the robustness of different methods to threshold selection, we introduce the second metric AuIoU. AuIoU provides a comprehensive measure of the model's accuracy in detecting OoD masks across various threshold levels, reflecting the ease of selecting the most suitable threshold. A higher AuIoU score indicates greater ease in selecting the optimal threshold. The third metric, mean F1 score, which takes into account both precision and recall, thus providing a more holistic assessment of the prediction results. Across all the three metrics, the proposed S2M outperforms the state-of-the-art OoD detection methods with a large margin.

\textbf{IoU} is a widely used evaluation metric in semantic segmentation. It is employed to assess the accuracy of the model in detecting OoD objects in comparison with the given labels. In this study, we ensure that for all methods which produce anomaly scores, the reported IoU represents the best IoU achieved by the optimal threshold on the specific dataset. For the proposed S2M, the reported IoU is calculated without the need for a threshold. During the computation process, we utilized all produced box prompts, obtaining the IoU by taking the intersection of the masks generated from these prompts. 

\begin{figure*}[tbp]
	\centering
        \scriptsize
	\begin{subfigure}{0.45\linewidth}
		\centering
		\includegraphics[width=1\linewidth]{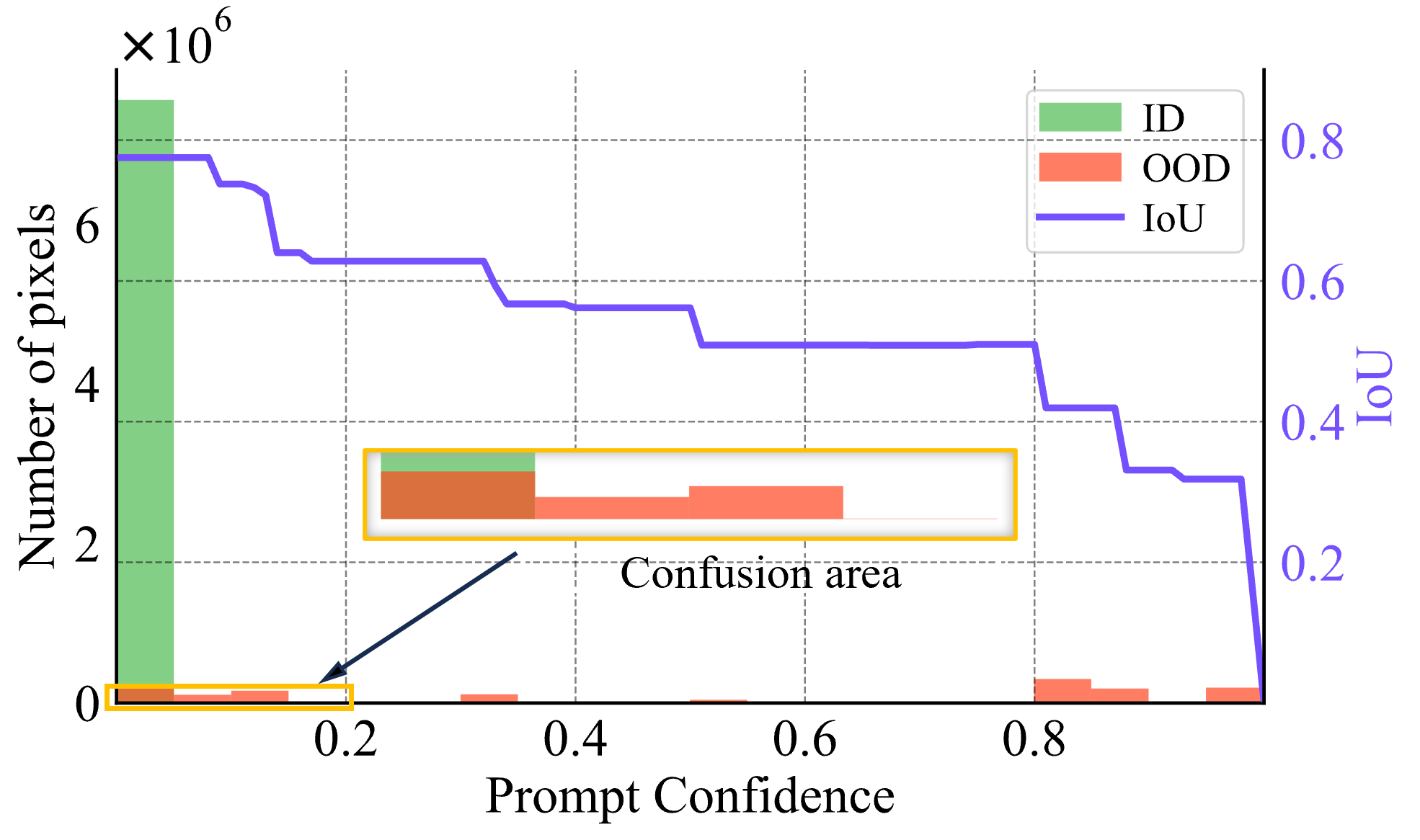}
		\caption{S2M}
		\label{test1_supp}
	\end{subfigure}
	\centering
	\begin{subfigure}{0.45\linewidth}
		\centering
		\includegraphics[width=1\linewidth]{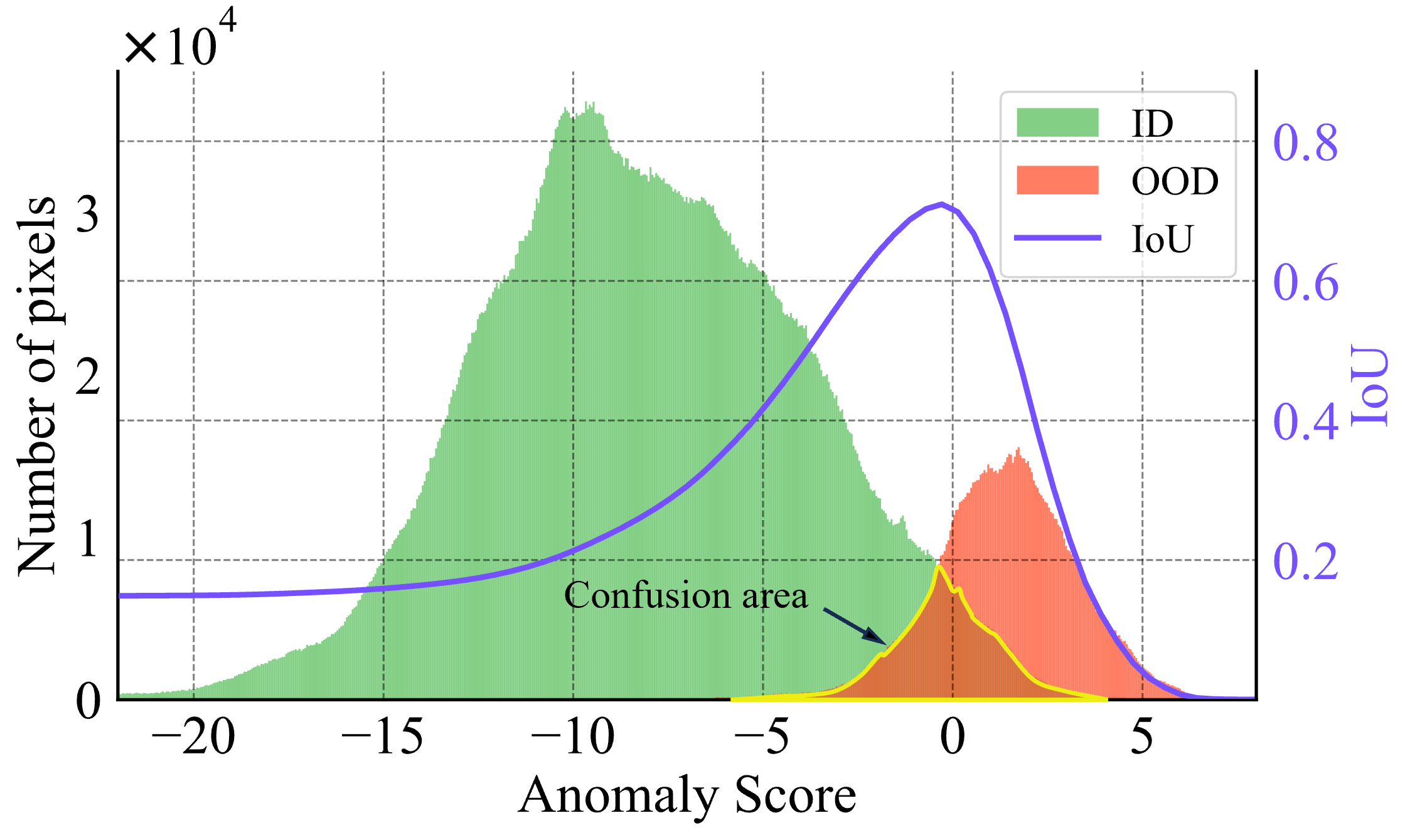}
		\caption{RPL}
		\label{test2_supp}
	\end{subfigure}
	\caption{\textbf{Anomaly score distribution and IoU curve.} We magnify the confusion area as depicted in Figure (a), to provide a clearer and more detailed view.}
	\label{largedistributon}
\end{figure*}

The average IoU of our S2M method is at least 7.52\% higher than the other methods listed in Table \ref{table1_compareSOTA}. This demonstrates that our method not only outperforms mainstream methods but also achieves superior performance without the necessity of a threshold. This result can be visualized in the Fig. \ref{largedistributon}. S2M achieves the highest IoU on the SMIYC validation dataset without a threshold. This indicates that our S2M method is more suitable for real-world application scenarios.

\textbf{AuIoU.} Area under IoU curve (AuIoU) is calculated by the area under the IoU curve with different thresholds. Here we define $th$ as the threshold. $TP_{th}$, $FP_{th}$, $FN_{th}$ represent the pixel numbers of True Positives, False Positives, and False Negatives when the threshold is $th$. True Positives (TP) are pixels correctly identified as OoD, False Positives (FP) are in-distribution pixels incorrectly identified as OoD, and False Negatives (FN) are OoD pixels that are not identified as such. With the above definitions, AuIoU can be computed as,
\begin{equation}
AuIoU = \frac{1}{n}  \sum_{th=th_{0}}^{th_{n}} \left (\frac{TP_{th}}{TP_{th}+FP_{th}+FN_{th}}\right )  
\end{equation}
where $n$ is the total number of steps and $th_0$ is the smallest threshold and $th_n$ is the largest threshold. In our experiments, we fixed the value of $n$ at 100, set $th_0$ to 0, and incrementally increased it to $th_n = 0.99$ with a step size of 0.01. A straightforward interpretation of AuIoU is the area under the IoU curve as depicted in Fig. \ref{largedistributon}. A higher AuIoU signifies that the model achieves better overall results across various thresholds, indicating that it is easier to obtain an appropriate threshold for the model. This is important in real-world application scenarios where determining the optimal threshold is inherently challenging. 

The average AuIoU of our S2M method, as shown in Table. \ref{table1_compareSOTA}, is 41.37\% higher than that of RPL. This suggests that RPL is sensitive to threshold selection. This perspective is also intuitively substantiated by observing the IoU curves in Fig. \ref{largedistributon}. The IoU curve for RPL shows that only a limited range of thresholds result in an IoU above 50\%, suggesting that RPL has a narrow range of thresholds where it can achieve optimal performance. This finding highlights the challenges RPL faces in determining an appropriate threshold for optimal performance, a significant limitation in practical applications where flexibility and adaptability in threshold settings are crucial. In contrast, our S2M method demonstrates superior performance in the accurate detection of OoD objects, working effectively without the need for threshold selection, in contrast to the limitations faced by RPL.

\textbf{Mean F1.}
The mean F1 score is calculated as the average of F1 scores obtained at various threshold levels. It is the harmonic mean of precision and recall, used to measure the accuracy and completeness of a model's predictions for the positive class. $Precision_{th}$ represent the precision when threshold is $th$. With the above definitions, mean F1 can be computed as,
\begin{equation}
\text{mean F1} =\frac{1}{n}  \sum_{th=th_{0}}^{th_{n}} \left (2\times \frac{Precision_{th}\times Recall_{th}}{Precision_{th}+Recall_{th}}\right )      
\end{equation}
This metric is especially valuable in scenarios where an optimal threshold has not been pre-established. A high F1 score indicates that the model achieves a favorable balance between precision and recall, suggesting it is proficient in correctly classifying positive cases while minimizing the number of false positives and false negatives. This implies the model's effectiveness in handling cases where both the accuracy of the positive predictions and the completeness of capturing all positive instances are critically important.

The average mean F1 of our S2M method on five datasets in Table \ref{fig:MainCompare} is 35.64\% higher than Synboost, which shows the best performance in mean F1 among mainstream methods. This indicates that our S2M method excels in balancing precision and recall, particularly in terms of accurately and comprehensively predicting positive classes. Specifically, the higher mean F1 score suggests that the S2M method is more effective in reducing both false positives (incorrectly marking negative instances as positive) and false negatives (missing true positive instances), thereby surpassing other mainstream methods in overall performance. This advantage is crucial as it demonstrates the reliability and accuracy of the S2M method across various application scenarios.

\subsection{Details of Efficiency Analysis}
\label{subsec:EA}
When comparing our S2M-B, used in our main experiments, with the RPL method, we observe that the total running time for S2M-B is only 0.059s longer than RPL, a modest increase considering its additional capabilities. The efficiency of S2M-B can be attributed to its dual-component structure which has shown in Fig. \ref{inference}. Firstly, it includes a mainstream OoD detector that generates an anomaly mask. Secondly, it features the SAM, which utilizes the original image and a box prompt to create precise OoD masks. A significant advantage of this setup is the efficiency in processing time. The operation of SAM on the image can be overlapped with the running time of the RPL, as these two processes can be executed in parallel. Once the box prompts are generated, they can be directly inputted into the decoder, together with the processed original image, to produce the final outcomes. Therefore, our method introduces minimal latency overhead compared to the baseline RPL.
\begin{table}[]
    \centering
    %\scriptsize
    \small
    \setlength{\tabcolsep}{1.5pt}
    \begin{tabular}{@{}ccccccc@{}}
    \toprule
    {}  & \multicolumn{3}{c}{{SMIYC anomaly}}  & \multicolumn{3}{c}{{SMIYC obstacle}}  \\ \cmidrule(l){2-4} \cmidrule(l){5-7}  
    {Method} & {AuIoU}& {IoU}   & {mean F1}       & {AuIoU} & {IoU}  & mean F1   \\ \cmidrule(r){1-7}
    {RPL+CoroCL}& {22.26}& {68.80}& {31.55}& {3.85}& {\textbf{28.66}}& {5.72} \\
    {S2M (FastSAM-s)}& {38.55}& {63.37}& {41.41}& {\textbf{21.66}}& {27.47}& {\textbf{25.00}} \\
    {S2M (FastSAM-x)}& {\textbf{60.58}}& {\textbf{81.09}}& {\textbf{64.18}}& {20.53}& {24.08}& {23.27} \\

    %{5\%} & {73.02} & {\textbf{70.81}}& {66.42} & {71.64} & {57.05} & {57.15}\\ \bottomrule 67.13 67.115
    \bottomrule
    \end{tabular}
    \caption{This table presents the accuracy measurements of S2M utilizes FastSAM on SMIYC. It highlights the S2M method with FastSAM, which utilizes anomaly scores from RPL+CoroCL as inputs. The parameters of FastSAM-x is 68M and FastSAM-s is 11M.}
    \label{fastsam}
\end{table}

\subsection{S2M with FastSAM}
\label{subsec:FastSAM}
As a faster version of SAM that performs comparably, FastSAM \cite{fastsam} can also be used as a promptable segmentation model in our S2M. FastSAM offers two unique model sizes: the compact and swifter FastSAM-s, based on YOLOv8s with an 11M model size, and the more extensive FastSAM-x, based on YOLOv8x with a 68M model size. We leverage FastSAM as the segmentation model and conduct experiments across all datasets using models trained with 2\% noise. From Table. \ref{fastsam}, we can find that the FastSAM also show an acceptable result on various metrics. S2M with FastSAM-x performs better on SMIYC anomaly validation dataset than RPL, with AuIoU 38.32\% higher, IoU 12.29\% higher and mean F1 32.63\% higher than RPL method. And the S2M with FastSAM-s performs better on SMIYC obstacle validation dataset, with AuIou 17.81\% higher and mean F1 19.28\% higher than RPL, but IoU 1.19\% lower than RPL. Here we use the IoU of RPL with the best performance on the validation dataset. The running time of S2M (FastSAM-s) and S2M (FastSAM-x) is shown in Table. \ref{fastSAM_runtime}. Due to the fast encoder speed and parallel way of segmentation model, the running time of S2M (FastSAM-s) and S2M (FastSAM-x) mainly influenced by the prompting process. However, the performance of FastSAM is lower than SAM with the same input. After visualization we found that SAM shows a stronger robustness to noisy box prompts than FastSAM. That is the reason that S2M with SAM performs better than FastSAM.

\begin{table}[!]
    \centering
    \scriptsize
    \begin{tabular}{@{}ccccc@{}}
    \toprule
    {Methods}&{RPL}&{S2M (FastSAM-s)} & {S2M (FastSAM-x)}\\
    \midrule
    {running time (\/s)} &0.2166&0.2415 & 0.2336\\
    \bottomrule
    \end{tabular}
    \caption{In these time measurements, we have excluded the dataloader aspect of the RPL model from our analysis, while including the set image process of the FastSAM model for consideration which run in a parallel way.} 
    \label{fastSAM_runtime}
\end{table}

\begin{table*}[!ht]
\small
\setlength{\tabcolsep}{1.5pt}
\begin{tabular*}{\linewidth}{@{\extracolsep{\fill}}r| ccc | ccc | ccc | ccc | cccc@{}}
\toprule
\multicolumn{1}{c}{\multirow{2}{*}{\makecell{Methods}} }
& \multicolumn{3}{c}{FS Static} & \multicolumn{3}{c}{FS Lost\&Found} & \multicolumn{3}{c}{SMIYC-Anomaly} & \multicolumn{3}{c}{SMIYC-Obstacle} & \multicolumn{3}{c}{RoadAnomaly} \\
\cmidrule(lr){2-4} \cmidrule(lr){5-7} \cmidrule(lr){8-10} \cmidrule(lr){11-13} \cmidrule(lr){14-16}
& \makecell{AuIoU} & \makecell{IoU} & \makecell{mean F1} & \makecell{AuIoU} & \makecell{IoU} & \makecell{mean F1} & \makecell{AuIoU} & \makecell{IoU} & \makecell{mean F1} & \makecell{AuIoU} & \makecell{IoU} & \makecell{mean F1} & \makecell{AuIoU} & \makecell{IoU} & \makecell{mean F1} \\
\midrule
% RPL (Entropy)& 24.26 & 50.12 & 32.87 & 9.90 & 23.31 & 14.40 & 31.09 & 48.85 & 41.93 & 18.23 & 36.33 & 24.60 & 16.66 & 26.23 & 25.10\\
% S2M (Entropy) & \textbf{76.05} & \textbf{84.65} & \textbf{79.95} & \textbf{32.12} & \textbf{26.16} & \textbf{38.60} & \textbf{37.80} & \textbf{58.13} & \textbf{47.57} & \textbf{59.36} & \textbf{58.48} & \textbf{65.03} & \textbf{25.40} & \textbf{30.83} & \textbf{31.56} \\
RPL (Entropy)& 8.81 & 14.65 & 14.61 & 2.25 & 3.63 & 3.83 & 30.03 & 47.02 & 40.96 & 0.72 & 1.45 & 1.35 & 16.66 & 26.23 & 25.10\\
S2M (Entropy) & \textbf{67.48} & \textbf{72.18} & \textbf{73.81} & \textbf{28.19} & \textbf{33.17} & \textbf{34.86} & \textbf{40.11} & \textbf{55.67} & \textbf{50.80} & \textbf{6.08} & \textbf{42.17} & \textbf{8.25} & \textbf{25.14} & \textbf{30.87} & \textbf{31.41} \\
\bottomrule
\end{tabular*}
\caption{RPL (Entropy) represent RPL+CoroCL methods with entropy-based anomaly score. S2M (Entropy) represent S2M with anomaly score from RPL (Entropy). The S2M method demonstrates strong \textbf{generalization capability}, effectively detecting OoD objects even when processing anomaly scores calculated using the entropy-based method.}
\label{entropy_score}
\end{table*}

\begin{table*}[!ht]
\small
\setlength{\tabcolsep}{1.5pt}
\begin{tabular*}{\linewidth}{@{\extracolsep{\fill}}r| ccc | ccc | ccc | ccc | cccc@{}}
\toprule
\multicolumn{1}{c}{\multirow{2}{*}{\makecell{Methods}} }
& \multicolumn{3}{c}{FS Static} & \multicolumn{3}{c}{FS Lost\&Found} & \multicolumn{3}{c}{SMIYC-Anomaly} & \multicolumn{3}{c}{SMIYC-Obstacle} & \multicolumn{3}{c}{RoadAnomaly} \\
\cmidrule(lr){2-4} \cmidrule(lr){5-7} \cmidrule(lr){8-10} \cmidrule(lr){11-13} \cmidrule(lr){14-16}
& \makecell{AuIoU} & \makecell{IoU} & \makecell{mean F1} & \makecell{AuIoU} & \makecell{IoU} & \makecell{mean F1} & \makecell{AuIoU} & \makecell{IoU} & \makecell{mean F1} & \makecell{AuIoU} & \makecell{IoU} & \makecell{mean F1} & \makecell{AuIoU} & \makecell{IoU} & \makecell{mean F1} \\
\midrule
% RPL (Entropy)& 24.26 & 50.12 & 32.87 & 9.90 & 23.31 & 14.40 & 31.09 & 48.85 & 41.93 & 18.23 & 36.33 & 24.60 & 16.66 & 26.23 & 25.10\\
% S2M (Entropy) & \textbf{76.05} & \textbf{84.65} & \textbf{79.95} & \textbf{32.12} & \textbf{26.16} & \textbf{38.60} & \textbf{37.80} & \textbf{58.13} & \textbf{47.57} & \textbf{59.36} & \textbf{58.48} & \textbf{65.03} & \textbf{25.40} & \textbf{30.83} & \textbf{31.56} \\
Mask2Anomaly& 6.00 & 11.60 & 10.57 & 0.68 & 1.57 & 1.29 & 44.08 & 81.31 & 53.37 & 8.07 & 42.41 & 11.90 & 24.31 & \textbf{56.74} & 32.80\\
Mask2Anomaly* & \textbf{57.77} & \textbf{62.34} & \textbf{62.90} & \textbf{25.74} & \textbf{27.94} & \textbf{30.94} & \textbf{65.61} & \textbf{83.90} & \textbf{72.69} & \textbf{54.69} & \textbf{60.58} & \textbf{61.46} & \textbf{47.18} & 53.39 & \textbf{52.56} \\
RbA& 3.28 & 8.07 & 5.99 & 0.45 & 1.52 & 0.86 & 28.03 & 70.28 & 39.07 & 2.49 & 23.18 & 4.22 & 16.43 & 54.99 & 24.32\\
RbA* & \textbf{50.41} & \textbf{57.16} & \textbf{56.66} & \textbf{24.46} & \textbf{27.27} & \textbf{29.12} & \textbf{33.33} & \textbf{76.23} & \textbf{41.48} & \textbf{41.99} & \textbf{52.09} & \textbf{48.40} & \textbf{44.13} & \textbf{55.36} & \textbf{51.30} \\
\bottomrule
\end{tabular*}
\caption{Mask2Anomaly* and RbA* denote the application of our S2M methodology utilizing the anomaly scores from Mask2Anomaly and RbA, respectively. Mask2Anomaly and RbA experiments is conducted with the models provided by authors. The results indicate that our method significantly improves the performance of stronger results.}
\label{table3_compareM2A_S2M}
\end{table*}

\begin{table}[!htb]
 \scriptsize
    \begin{center}
    \setlength{\tabcolsep}{1.8pt}
        \begin{tabular}{c c c c c c c}
        \toprule
            & \multicolumn{3}{c}{SMIYC Anomaly}&\multicolumn{3}{c}{SMIYC Obstacle}\\ 
            \hline
            %\cmidrule(lr){2-3} \cmidrule(lr){4-5} \cmidrule(lr){6-7}
            Method& AuIoU& IoU& meanF1& AuIoU& IoU& meanF1\\
            vanilla RPL
& 22.26& 68.80& 31.55& 3.85&  28.66& 5.72\\
            S2M (RPL w. PEBAL's generator)  & \textbf{59.78}& \textbf{73.85}& \textbf{68.41}&  \textbf{13.96}& \textbf{29.41}&  \textbf{20.24}
\\\bottomrule
        \end{tabular}
    \end{center}
    \caption{Compare the results of RPL and SAM with prompt generator trained on anomaly score from Pebal.}
    \label{reuse}
\end{table}

\subsection{S2M With Entropy Based Anomaly Score} 
\label{subsec:EB}
The anomaly scores in our methods, derived from RPL, have been computed using an energy-based approach. To demonstrate the generalization capability of our method, we have also conducted experiments using anomaly scores calculated via an entropy-based method \cite{chan2021entropy}. As previously mentioned, we employ RPL \cite{RPL} to generate anomaly scores for training images using an entropy-based method, while maintaining all other settings unchanged. Given that the range of entropy-based anomaly scores approximately lies between 0 and 1, we amplify the anomaly score of each pixel by a factor of 20 during training and inference to facilitate the model's ability to distinguish between in-distribution and out-of-distribution pixels. The results of entropy-based anomaly score of RPL and S2M based entropy anomaly score are shown in Table \ref{entropy_score}. The table demonstrates that S2M, when utilizing anomaly scores calculated via the entropy-based method, also exhibits improved performance compared to using the original anomaly scores.

\subsection{S2M with Advanced SOTA Methods}
\label{subsec:S2M With Advanced SOTA Methods}

To demonstrate the general improvement capability of our method, we conducted experiments using the anomaly scores from Mask2Anomaly~\cite{m2a} and RbA~\cite{RbA}. These experiments were carried out with the models provided by the authors, applying our S2M based on their anomaly scores. The results, shown in Table \ref{table3_compareM2A_S2M}, indicate that our method consistently enhances the performance. 

\subsection{Reuse Prompt Generator without Training}
\label{subsec:Reuse prompt generator without training}
Results presented in Table \ref{reuse} demonstrate that a prompt generator, when trained on PEBAL's \cite{PEBAL} anomaly scores and evaluated on RPL' \cite{RPL}, still achieves superior performance compared to RPL. The initial training of PEBAL's generator utilized anomaly scores from PEBAL, which has different domain of anomaly score from RPL. Applying PEBAL's generator directly on RPL's anomaly scores, without any modification, typically yields suboptimal results. In our experiment, we scale up the anomaly score of RPL by a factor of 20. This adjustment contributes to better performance. The result suggests that our prompt generator can be effectively used without the need for retraining.

\subsection{Input Contains No OoD Objects}
\label{subsec:Input Contains No OoD Objects}
We test our S2M (RPL) method on Cityscapes validation datasets, which comprises 600 images without OoD objects and used as ID dataset during training.  The result shows that our prompt generator did not detect any box prompt in all 600 images, indicating that S2M can effectively discern images without OoD objects. 

\subsection{Visualizations of Segmentation Result}
\label{subsec:Visualization}
\begin{figure*}[p]
    \centering
    \includegraphics[width=1\textwidth]{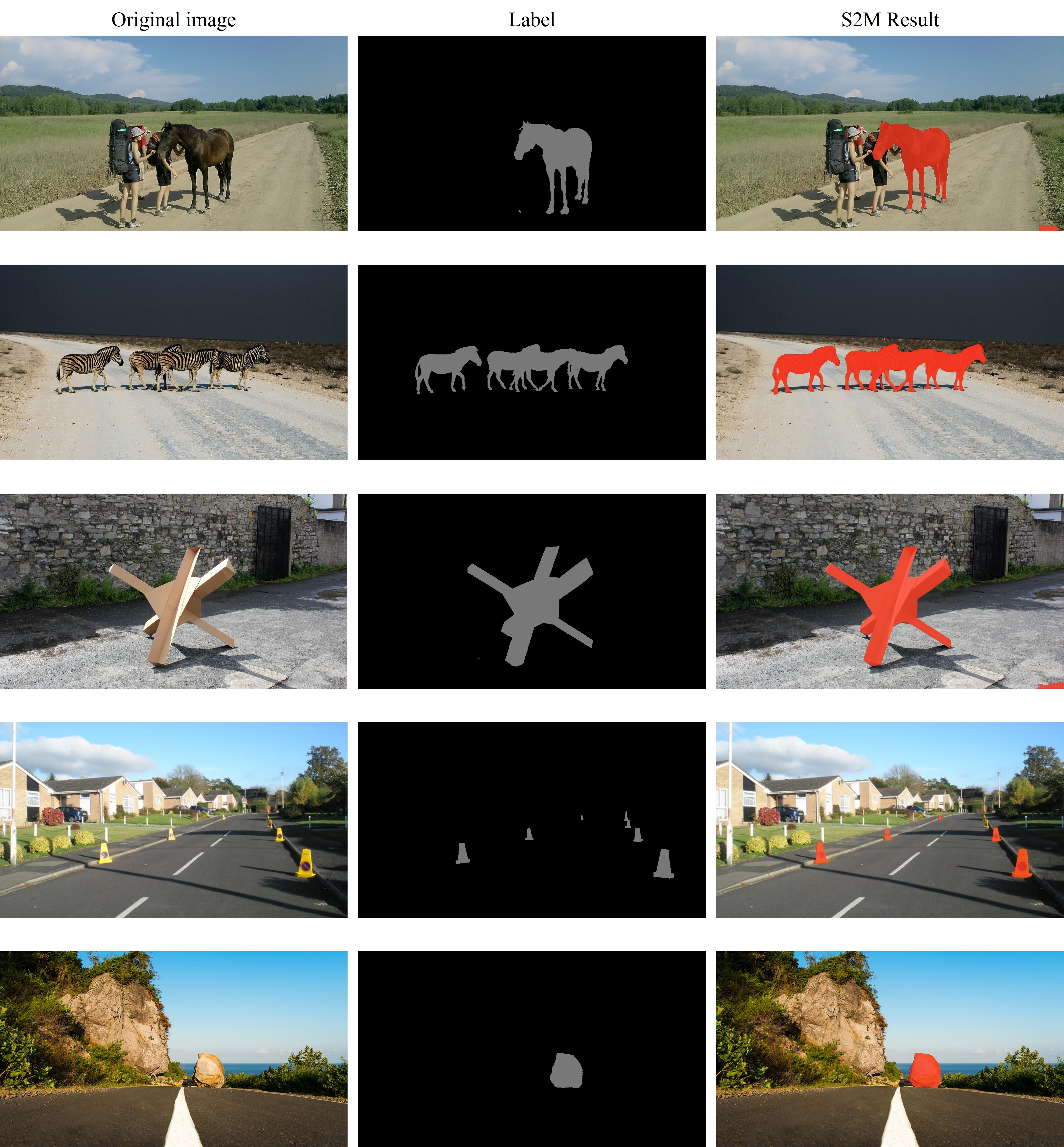}
    \caption{Visualization of S2M on Road Anomaly validation set. In the annotated images, pixels colored gray represent OoD objects, black pixels denote ID objects.}
    \label{RoadAnomaly}
\end{figure*}

\begin{figure*}[p]
    \centering
    \includegraphics[width=1\textwidth]{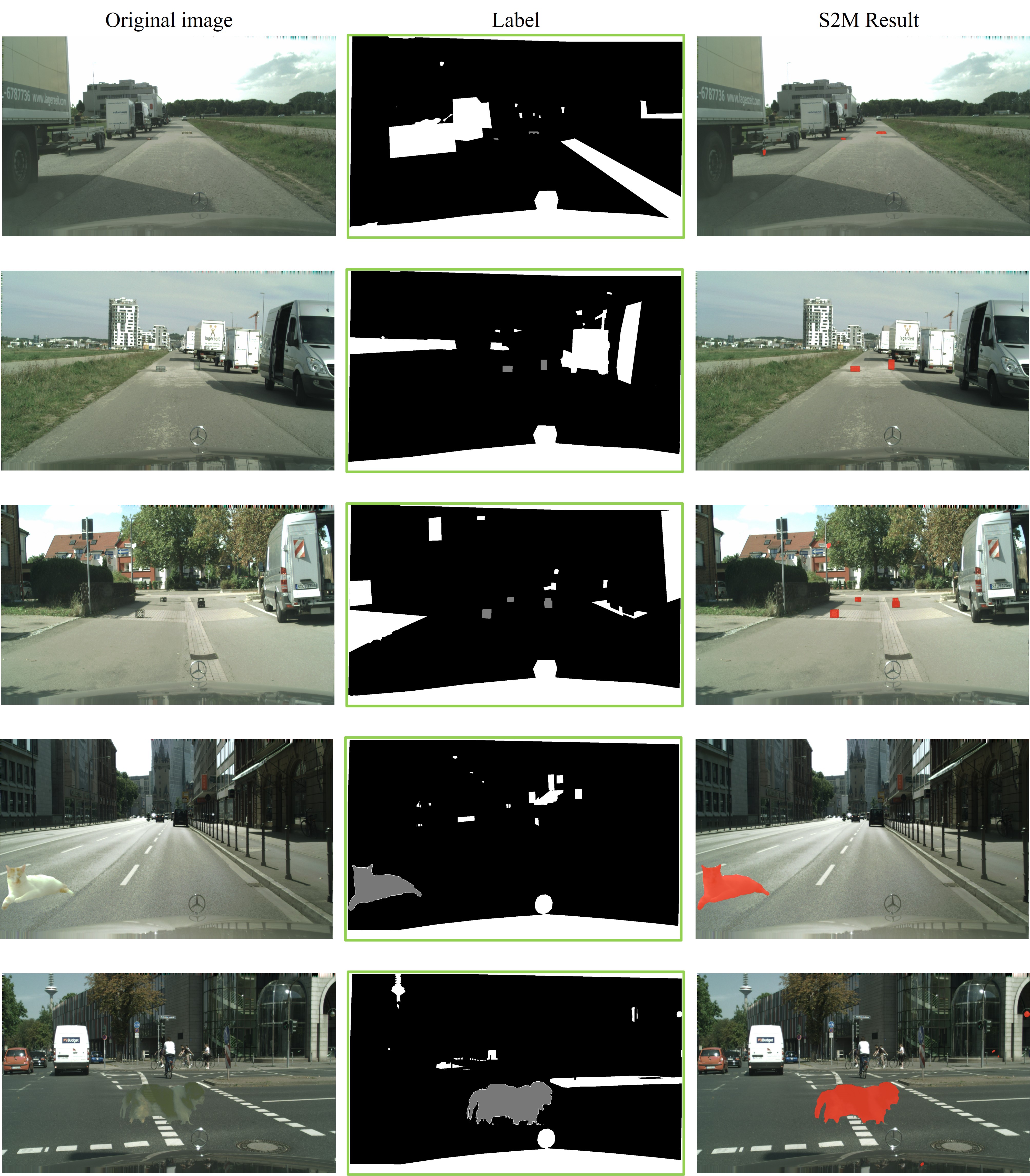}
    \caption{Visualization of S2M on Fishyscapes validation set. In the annotated images, pixels colored gray represent OoD objects, black pixels denote ID objects, and white pixels indicate regions to be ignored.}
    \label{Fishyscapes}
\end{figure*}

\begin{figure*}[p]
    \centering
    \includegraphics[width=1\textwidth]{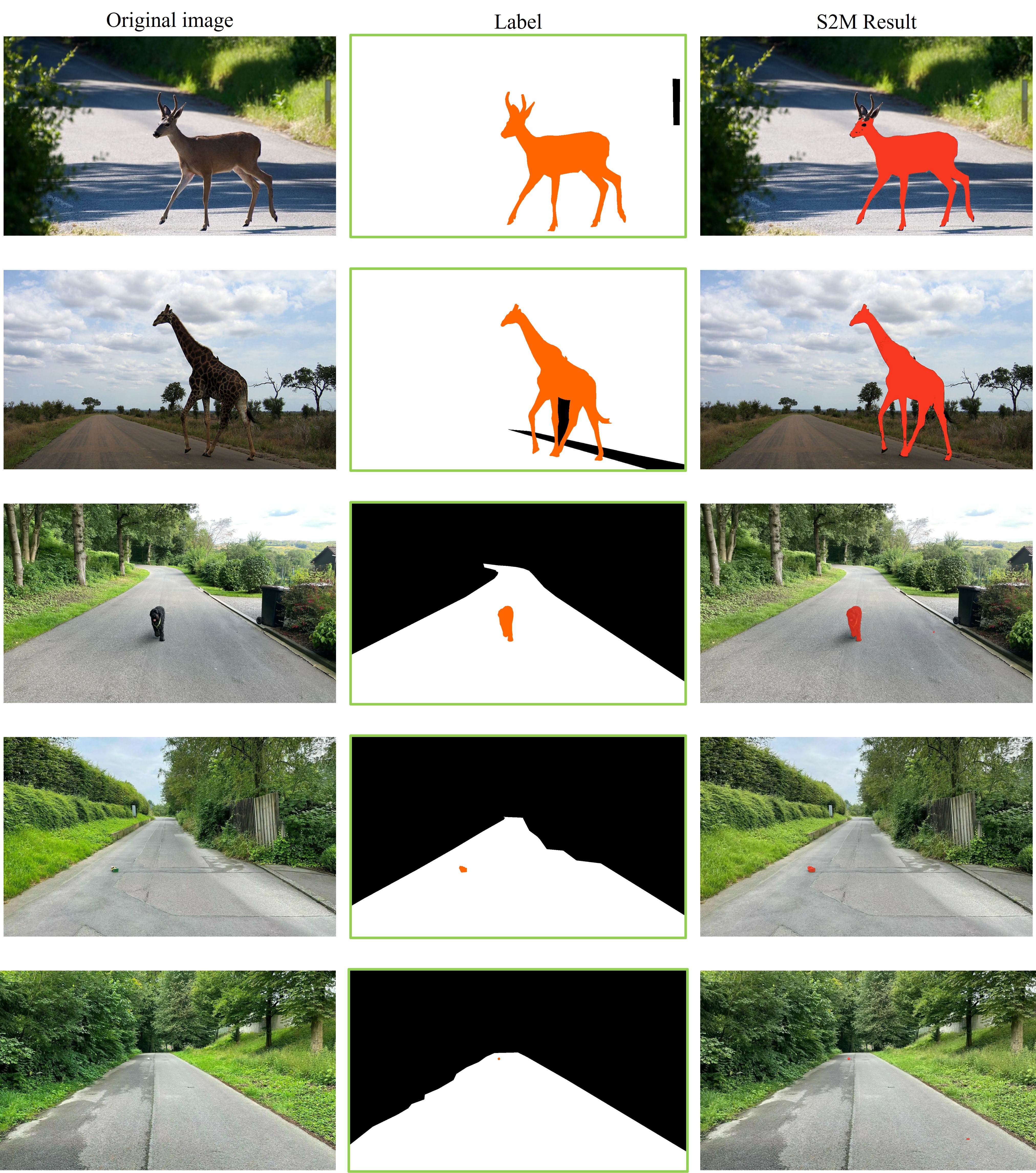}
    \caption{Visualization of S2M on SMIYC validation set. In the annotated images, pixels colored orange represent OoD objects, white pixels denote ID objects, and black pixels indicate regions to be ignored.}
    \label{SMIYC}
\end{figure*}

We visualize the OoD mask generated by our S2M methods on Road Anomaly, Fishyscapes and SMIYC in Fig. \ref{RoadAnomaly}, Fig. \ref{Fishyscapes} and Fig. \ref{SMIYC}. 

Validation on Road Anomaly demonstrates the precision of S2M. Our method accurately detects OoD objects while ensuring that ID objects are not mistakenly identified as OoD, as shown in the first row of Fig. \ref{RoadAnomaly}. S2M gives a precise mask of horse and excludes the people nearby. S2M is also capable of generating precise masks for multiple OoD objects, as demonstrated in the second and fourth rows.

Validation on the Fishyscapes dataset highlights the precision of S2M in detecting small anomalies. Our method excels in accurately identifying small OoD objects when the anomaly scores are optimal, as illustrated in the first row of Fig. \ref{Fishyscapes}. This capability is crucial for scenarios involving diminutive and subtle anomalies. Furthermore, S2M efficiently detects semi-transparent, synthetically created OoD objects, showcasing its robustness and precision in complex scenarios. This is effectively demonstrated in the fourth and fifth rows, where S2M successfully delineates these challenging objects without compromising accuracy.

The SMIYC dataset exemplifies the efficacy of our approach in addressing the diverse and dynamic nature of road obstacles. The comprehensive environment of SMIYC allows for the evaluation of our method's ability to detect a wide range of OoD objects on roadways, from tiny to larger, more conspicuous obstacles.

\end{document}